\documentclass{article}

 \usepackage[preprint]{neurips_2025}

\usepackage{authblk}

\usepackage[utf8]{inputenc}    % Allow utf-8 input (for pdfLaTeX)
\usepackage[T1]{fontenc}       % Use 8-bit T1 fonts
\usepackage{microtype}         % Improved typography (load only once)

\usepackage{amsmath}           % Advanced math environments
\usepackage{amsfonts}          % Blackboard math symbols (e.g., \mathbb)
\usepackage{nicefrac}          % Compact symbols for 1/2, etc.

\usepackage{tabularx,booktabs,array} % Professional-quality tables (booktabs loaded here)
\usepackage{caption}
\renewcommand{\arraystretch}{1.15}   % Increase row height in tables

\usepackage{graphicx}
\usepackage{subcaption}
\usepackage{xcolor}
\usepackage[most]{tcolorbox}

\usepackage{enumitem}
\setlist[itemize]{nosep,leftmargin=1.2em} 
\usepackage{csquotes}
\MakeOuterQuote{"}
\tcbset{
  promptbox/.style={
    enhanced, breakable,
    colback=white,
    colframe=blue!70!black,  % Using blue as an example frame color
    left=2.5mm, right=2.5mm, top=2mm, bottom=2mm,
    boxsep=3.5mm, arc=2pt, outer arc=2pt,
    before skip=6pt, after skip=6pt
  }
}

\usepackage{afterpage}
\usepackage[colorinlistoftodos,textsize=footnotesize]{todonotes}

\usepackage{hyperref}          % For clickable links and references
\title{Another Turn, Better Output? \raisebox{-0.4em}{\includegraphics[height =1.9em]{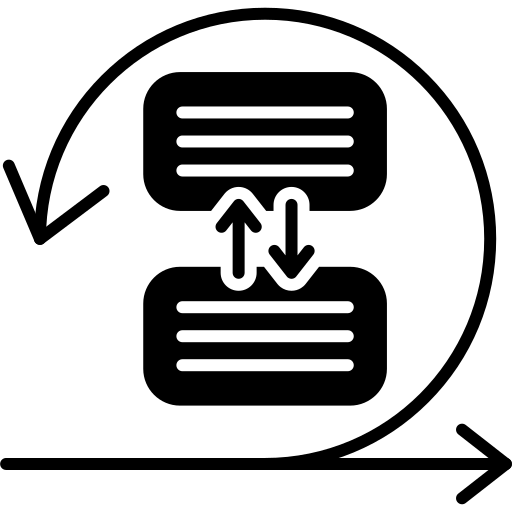}}:  

A Turn-Wise Analysis of Iterative LLM Prompting}
\usepackage{graphicx} % enables \includegraphics
\graphicspath{{figs/}} % optional: folder where your PDFs live

\author[1]{\textbf{Shashidhar Reddy Javaji}}
\author[2]{\textbf{Bhavul Gauri}}
\author[1]{\textbf{Zining Zhu}}

\affil[1]{Stevens Institute of Technology}
\affil[2]{Meta}

\affil[ ]{\texttt{sjavaji@stevens.edu, bhavul@meta.com, zzhu41@stevens.edu}}

\begin{document}

\maketitle

\begin{abstract}

Large language models (LLMs) are now used in multi-turn workflows, but we still lack a clear way to measure when iteration helps and when it hurts. We present an evaluation framework for iterative refinement that spans ideation, code, and math. Our protocol runs controlled 12-turn conversations per task, utilizing a variety of prompts ranging from vague ``improve it'' feedback to targeted steering, and logs per-turn outputs. We score outcomes with domain-appropriate checks (unit tests for code; answer-equivalence plus reasoning-soundness for math; originality and feasibility for ideation) and track turn-level behavior with three families of metrics: semantic movement across turns, turn-to-turn change, and output size growth.
Across models and tasks, gains are domain-dependent: they arrive early in ideas and code, but in math late turns matter when guided by elaboration. After the first few turns, vague feedback often plateaus or reverses correctness, while targeted prompts reliably shift the intended quality axis (novelty vs. feasibility in ideation; speed vs. readability in code; in math, elaboration outperforms exploration and drives late-turn gains). We also observe consistent domain patterns: ideation moves more in meaning across turns, code tends to grow in size with little semantic change, and math starts fixed but can break that path with late, elaborative iteration.Together, the framework and metrics make iteration measurable and comparable across models, and signal when to steer, stop, or switch strategies.
\end{abstract}

\section{Introduction}

The advent of Large Language Models (LLMs) has shifted the paradigm of human-computer interaction, moving beyond one-shot prompting to more dynamic, multi-turn workflows \citep{sahoo_systematic_2025, li_beyond_2025}. Foundational to this shift is instruction tuning, which aligns models to follow human feedback and engage in collaborative tasks \citep{ouyang_training_2022}. Central to this new paradigm is the process of iterative refinement, where a user and an AI progressively improve an initial output \citep{xue_improve:_2025}. This approach has become a key of modern LLM applications, with frameworks like SELF-REFINE and Reflexion demonstrating that models can even use self-generated feedback to enhance their outputs, highlighting the immense potential of iterative loops \citep{madaan_self-refine:_2023, shinn_reflexion:_2023}. This evolution towards multi-step interaction leverages emergent cognitive capabilities of modern LLMs, which can manifest without explicit prompting \citep{arnold_phase_2024}.
To enhance model reasoning and performance, a significant body of research has focused on developing sophisticated, structured prompting techniques. Seminal approaches like Chain-of-Thought (CoT), which breaks down problems into intermediate steps, have been shown to elicit stronger reasoning \citep{wei_chain_2022}. This has been extended to more complex methods like Tree-of-Thoughts (ToT), which explores multiple reasoning paths, and ReAct, which synergizes reasoning with actions \citep{yao_tree_2023}. Other methods focus on providing models with specific, domain-relevant knowledge to improve the quality of specialized code generation \citep{gu_effectiveness_2025} or employ complex search algorithms and multi-agent reflection to reinforce logical steps \citep{yuan_reinforce_2025}. These structured approaches have proven effective, demonstrating that with the right guidance, LLMs can be powerful and reliable reasoning engines.

However, a critical gap exists between these highly-structured, engineered prompting methods and the far more common, ``naive'' interaction style where users provide simple, vague feedback. The behavior of LLMs in these unguided, multi-turn loops is poorly understood, with studies showing performance can drop significantly in multi-turn conversations compared to single-turn tasks \citep{laban_llms_2025}. Recent work suggests this process is fraught with risk; for example, a simple iterative prompt like ``Are you sure'' can paradoxically decrease a model’s truthfulness and increase overconfidence \citep{krishna_understanding_2024}. This aligns with findings that models’ self-correction abilities are often brittle and unreliable \citep{ji_survey_2023}. This degradation may be exacerbated in long contexts, where models struggle to access information from the ``middle'' of the prompt history \citep{liu_lost_2024}. This creates a dangerous scenario analogous to ``model collapse'' or a game of ``broken telephone,'' where a system feeding on its own output can enter a degenerative cycle \citep{mohamed_llm_2025}. These gaps motivate three questions that guide our study: How sensitive are iterative refinements to ``word choice'' and instruction specificity? When does iterative refinement help---and when does it drift or collapse? If the do collapse, do all models collapse similarly?

We address these questions with a controlled, turn-by-turn study. We run 12-turn conversations across ideation, mathematical reasoning, and code, log every turn, and compare two feedback settings: (i) vague prompts using three near-synonyms (``improve,'' ``make it better,'' ``refine'') and (ii) specific steering along domain axes (novelty vs. practicality for ideation, speed vs. readability for code, elaboration vs. alternate method for math). Our evaluation focuses on dynamics, not just single-shot quality: we track innovation vs. stability across turns, growth in complexity and its plateaus, and semantic drift from the initial intent. We add a compact codebook of common failures (stagnation, over-engineering, flawed anchoring) and use LLM-assisted judgments. We also connect these dynamics to regime shifts suggested by recent work on phase-transition-style effects \citep{arnold_phase_2024, nakaishi_critical_2024}.

Our results show clear patterns. In ideas and code, when iteration helps, it does so early; in math, late turns can help when the prompt asks for elaboration. After a few turns, vague feedback often plateaus or reduces quality, while targeted steering reliably shifts the intended axis without large side effects. The degradation appears in domain-specific ways: ideation tends to repeat itself, code grows in size without meaningful change, math is fixed by default but can be broken by elaboration to find correct paths late. We quantify these effects with simple, defensible metrics---including Lexical Novelty (LN), growth factor, drift from origin, and turn-to-turn volatility---and we analyze sensitivity to word choice and instruction specificity across models.As context, SELF-REFINE exemplifies structured critique-and-revise loops where explicit guidance improves outcomes \citep{madaan_self-refine:_2023}. Finally, we translate these measurements into practical guidance: when to steer with concrete goals, when to stop to avoid harm, and when to switch strategies to limit semantic drift and related risks \citep{spataru_know_2024}. Together, the framework, metrics, and protocol provide a reproducible basis for comparing models, prompts, and domains, and set up the rest of the paper’s methodology, evaluation, and results.

% \textbf{Research Questions:}

% \begin{itemize}
%     \item  How sensitive are iterative refinements to “word choice” and instruction specificity?
%     \item When does iterative refinement help—and when does it drift or collapse?
%     % \item  Do cross-model pipelines stabilize refinement?
%     \item Are ideation, math reasoning, and code synthesis differently susceptible to drift/collapse?
%     \item Does all models collapse similarly.
% \end{itemize}

\section{Related Work}

Multi-turn performance is consistently harder than single-turn prompting: large simulations show an average 39\% drop in multi-turn vs.\ single-turn across six generation tasks, driven chiefly by unreliability rather than aptitude loss \cite{laban_llms_2025}. Complementing this, \emph{MultiChallenge} finds that frontier models score <50\% (e.g., 41.4\% for Claude 3.5 Sonnet) on realistic multi-turn dialog despite strong results elsewhere \cite{sirdeshmukh2025multichallengerealisticmultiturnconversation}. On the remedy side, \emph{Self-Refine} improves initial outputs via self-feedback with 5--40\% absolute gains across tasks \cite{madaan_self-refine:_2023}; post-training that explicitly \emph{stimulates} self-refinement (e.g., \emph{ARIES}) reports strong improvements on AlpacaEval2/Arena-Hard and math benchmarks by iteratively collecting refinement data and optimizing preferences \cite{zeng2025evolvingllmsselfrefinementcapability}. Multi-agent, coarse-to-fine frameworks such as \emph{MAgICoRe} directly target excessive refinement and error-localization issues and outperform Self-Refine/Best-of/Self-Consistency while continuing to improve with more iterations \cite{chen2024magicoremultiagentiterativecoarsetofine}. Beyond optimization, clarifying-question policies (\emph{ACT}) use contrastive preference tuning to improve mixed-initiative multi-turn interactions \cite{chen2025learning}.

However, unguided iteration can harm truthfulness and calibration: asking models to re-check themselves (``Are you sure?''-style prompts) reduces accuracy and worsens calibration \cite{krishna_understanding_2024}; Some LLM-as-judge setups overstate confidence \cite{tian2025overconfidencellmasajudgediagnosisconfidencedriven}; here we use ground-truth checks where possible and report relative scores . Broader calibration studies document persistent miscalibration across sizes and settings (\emph{Mind the Confidence Gap}) \cite{chhikara2025mindconfidencegapoverconfidence}, while RLHF can induce \emph{verbalized} overconfidence; reward-calibrated RLHF mitigates this without hurting quality \cite{leng2025tamingoverconfidencellmsreward}. Data-feedback loops introduce additional risks: recursively training on model-generated data causes model collapse \cite{shumailov2024ai}, though accumulating synthetic \emph{with} real data can avoid collapse \cite{gerstgrasser2024is}; theory further shows collapse can still occur under certain conditions even with accumulation \cite{barzilai2025modelsdontcollapseconsistency}. Surveys synthesize multi-turn agent evaluation (nearly 250 sources) \cite{guan2025evaluatingllmbasedagentsmultiturn}, and recent work on \emph{self-iterative label refinement} proposes robust unlabeled-learning pipelines to denoise pseudo-labels, offering a safer iteration template for classification tasks \cite{asano2025selfiterativelabelrefinement}.
\section{Methodology}

\subsection{Task Domains and Dataset Curation}
To test our hypotheses across diverse cognitive tasks, we curated datasets of 50 problems each from three established benchmarks. For open-ended ideation, we sampled scientific idea generation tasks from LiveIdeaBench \citep{ruan2025liveideabenchevaluatingllmsdivergent} using a stratified approach for balanced domain representation. For structured code generation, we curated coding problems from DS-1000 \citep{lai2022ds1000naturalreliablebenchmark} with a quota-based strategy mirroring its distribution across key Python libraries. Finally, for formal mathematical reasoning, we assembled high-difficulty problems from Omni-MATH \citep{gao2024omnimathuniversalolympiadlevel} by filtering for tasks with a difficulty rating greater than 7/10, ensuring each requires sophisticated, multi-step reasoning.

\subsection{The Iterative Refinement Protocol}

Our experiment uses an automated multi-turn protocol that simulates a human--AI refinement loop. For each task, we use a fixed 12-turn ``conversation,'' aligning with community practice around \textasciitilde10 turns---MultiChallenge builds histories of up to 10 turns and MT-Eval structures dialogues with ten turns---while adding two extra iterations to capture late-stage changes.
\citep{deshpande-etal-2025-multichallenge}\citep{kwan-etal-2024-mt}

\textbf{Initial Generation (Turn 1):} The model is given the initial task prompt and generates its first response.

\textbf{Iterative Feedback (Turns 2-12):} For each subsequent turn, the model is presented with only its \textit{own output from the previous turn}, followed by a simple instruction to improve it.

This \textit{memoryless} protocol is intentionally designed to stress-test the model's internal coherence and ability to improve without being constantly re-grounded by the original prompt. To ensure statistical robustness, each task-model-prompt combination is run independently. The entire process is managed by the automated experimental runner script detailed in \autoref{prompts}.

\subsection{ Prompting Strategies}

A key component of our research is to understand how the \textit{nature} of the feedback influences the refinement trajectory. To this end, we designed two main experimental groups: \textbf{Vague Feedback} and \textbf{Specific Steering}.

    \textbf{The Vague Feedback Group:} This condition tests the model's "default" behavior. We test three semantically similar prompts to ensure our findings are robust to minor wording changes. This allows us to test whether the model's behavior is tied to the specific verb "improve" or the general semantic concept of improvement.
    \begin{itemize}
        \item \textbf{V1 (Baseline):} "This [idea/code/solution] can be better. \textbf{Improve it.}"
        \item \textbf{V2 (Synonym):} "This [idea/code/solution] can be better. \textbf{Make it better.}"
        \item \textbf{V3 (Refinement-Connotation):} "This [idea/code/solution] can be better. \textbf{Refine it.}"
    \end{itemize}

     \textbf{The Specific Steering Group:} This condition serves as a control, testing how models respond to clear, expert-like guidance. The prompts were chosen to represent fundamental, often opposing, goals within each domain.
    \begin{itemize}
        \item \textbf{For Ideation,} the prompts test the trade-off between innovation and applicability. We steer towards \textbf{novelty} ("Make this idea more novel and surprising") and \textbf{practicality} ("Make this idea more practical and feasible").
        \item \textbf{For Coding,} the prompts test a classic software engineering trade-off. We steer towards \textbf{performance} ("Refactor...for maximum execution speed") and \textbf{maintainability} ("Refactor...for maximum readability and clarity").
\item \textbf{For Math,} the prompts target the observed failure mode of "flawed anchoring." We steer towards \textbf{elaboration} ("Elaborate on each step with more detail") to test justification ability, and towards \textbf{exploration} ("Provide an alternative method...") to test flexibility.
    \end{itemize}
% \end{itemize}

\begin{figure}[t]
  \centering
  % single-page PDF, or choose a page from a multi-page PDF:
  \includegraphics[page=1,trim=62mm 25mm 62mm 16mm,clip,width=0.89\linewidth]{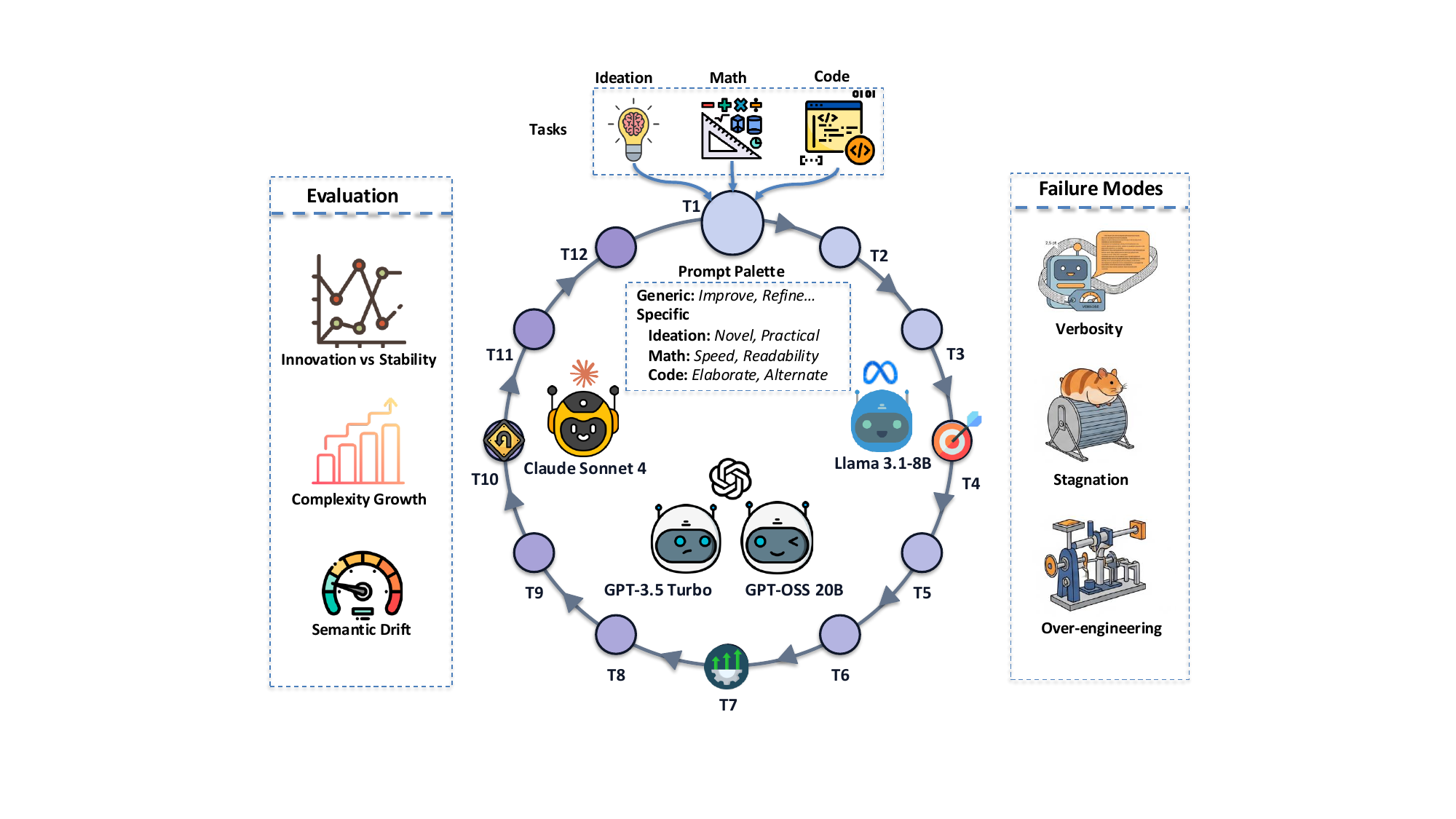}
  \caption{An overview of our experimental framework for studying iterative LLM refinement. We test four leading models across three distinct cognitive domains (Ideation, Math, and Code) using two primary feedback styles: vague prompts (e.g., "Improve it") and specific, expert-like prompts. We analyze the resulting multi-turn conversations using a suite of quantitative and qualitative metrics to identify behavioral "fingerprints," such as the tendency for an idea to stagnate, increase in complexity, or drift from its original intent.} 
\end{figure}

\subsection{Models }

To ensure our findings are generalizable, we conducted our experiments across four powerful, state-of-the-art LLMs representing diverse architectures and training methodologies. The specific models studied were:\textit{ GPT-3.5-Turbo \citep{ye2023comprehensivecapabilityanalysisgpt3}, Claude-Sonnet-4.0 \citep{anthropic_claude4_system_card_2025}, Llama-3.1-8B-Instruct \citep{grattafiori2024llama3}, GPT-OSS-20B \citep{openai_gpt_oss_model_card_2025}}. All models were accessed via their standard APIs or a local Hugging Face pipeline. The temperature was set to a consistent value of 0.7 and \emph{max\_tokens} of 10K across all experiments to balance creative, diverse outputs with a reasonable degree of coherence and reproducibility.

\subsection{Evaluation Framework}
To analyze the multi-turn outputs from our experiment, we developed a multi-faceted evaluation framework. Our approach is designed to be robust, largely automated, and capable of capturing the nuanced behaviors we observed in preliminary studies. The framework combines objective, ground-truth-based metrics, a suite of behavioral metrics to characterize the dynamics of the iterative process, and a scalable protocol for assessing semantic quality.

\subsubsection{Outcome \& Efficiency Metrics}
For the Math and Coding domains, we measure objective performance by assessing correctness at each turn. This allows us to understand the dynamics of success and failure throughout the iterative process. We calculate a binary Correctness Score for each of the 12 turns in every run. For \textbf{Coding} tasks (\texttt{DS-1000}), each code snippet is executed in a sandboxed environment and validated against the benchmark's unit tests.For \textbf{Mathematical Reasoning} (\texttt{OmniMath}), we evaluate each problem by sending the 12-turn JSON together with the ground-truth solution/answer to a Gemini model, which returns per-turn scores: a binary \emph{answer\_correctness} (final-answer equivalence) and a 1--10 \emph{reasoning\_soundness}.

\subsubsection{Behavioral Dynamics Metrics}
% This suite of metrics is designed to quantitatively characterize the ``Converge-Drift-Collapse'' (CDC) trajectory. We track three key behavioral signals: the conceptual path, the rate of new information, and the growth in output size.

\paragraph{Semantic Dynamics: Drift and Volatility,}
To characterize the dynamics of conceptual change, we track two metrics. First, we measure \textbf{Drift from Origin}, which quantifies how far the current idea has semantically strayed from its starting point. A higher score indicates greater drift. Second, we measure \textbf{Turn-to-Turn Volatility}, which captures the magnitude of change between consecutive turns. Letting \(V(t)\) be the sentence-embedding vector for the response at turn \(t\), the metrics are formally defined as:
\[
\text{Drift\_from\_Origin}(t) = 1 - \frac{V(1) \cdot V(t)}{\|V(1)\| \|V(t)\|}
\qquad
\text{Volatility}(t) = 1 - \frac{V(t-1) \cdot V(t)}{\|V(t-1)\| \|V(t)\|}
\]
% Plotting the Drift from Origin reveals the full trajectory of conceptual coherence, while a high Volatility score indicates a radical, unstable pivot.

All semantic similarity metrics, including Drift from Origin and Turn-to-Turn Volatility, were calculated using the \texttt{Qwen/Qwen3-Embedding-0.6B} \citep{qwen3embedding} sentence transformer model. This model was selected for its excellent balance of performance and efficiency. At the time of our experiments, it held a top-4 position on the Massive Text Embedding Benchmark (MTEB) leaderboard \citep{muennighoff-etal-2023-mteb}.

\paragraph{Lexical Novelty\citep{li-etal-2016-diversity}:}
To measure a model's ``creative stamina'' and pinpoint when it collapses into repetition, we track its \textbf{Lexical Novelty (LN)} at each turn. We define LN as the percentage of new phrases in a response that have not appeared in prior turns of the conversation. To capture both short and long repeated phrases, we use a combination of 2-grams (bigrams) and 3-grams (trigrams).

\paragraph{Growth Factor \citep{laban_llms_2025}:}
This metric quantifies the ``over-engineering'' and ``verbosity'' phenomena by tracking the turn-by-turn growth of the output. We first define a domain-specific \textbf{Growth Score}, \(G(t)\), for each turn: for \textbf{Ideation} and \textbf{Math}, it is the total word count; for \textbf{Coding}, it is the number of lines of code (LoC). The Growth Factor at turn \(t\) is then calculated by normalizing the current turn's score by the score of the initial turn.
% \[
% \text{Growth\_Factor}(t) = \frac{G(t)}{G(t=1)}
% \]

\subsubsection{Semantic Quality Metrics (LLM-as-a-Judge)}
To assess nuanced qualities, we employ a state-of-the-art LLM (Gemini 2.5-pro) as a scalable proxy for expert human judgment. Our protocol involves providing the evaluator LLM with a domain-specific scorecard to rate each turn's output, yielding a quality score.
For Ideation, we measure Feasibility and Novelty.
  For Coding, we measure Pragmatism (is the code appropriately scaled to the problem?) and Readability. For Math, we measure Logical Soundness and Clarity of Explanation.
To assess nuanced qualities that are difficult to measure automatically, we employ a state-of-the-art LLM as a scalable proxy for expert human judgment. Our approach is grounded in the ``LLM-as-a-Judge'' paradigm, which has been shown to achieve agreement with human expert annotators that is comparable to human-to-human agreement levels \citep{zheng2023judging}. For our evaluator, we selected \textbf{Gemini 2.5 Pro} \citep{google_gemini_25_pro_model_card_2025}, a top-performing model that, at the time of our experiments, holds a top-2 ranking on the Chatbot Arena leaderboard---a large-scale benchmark that operationalizes these evaluation principles.

% The validity of this method was confirmed via a study where the LLM-judge's scores showed a high Spearman's rank correlation \_\_\_\_\todo{work to be done} with human expert ratings on a representative subset of our data.

\section{Results}
% In the document
\begin{figure}[t]
  \centering

  \includegraphics[width=0.97\linewidth]{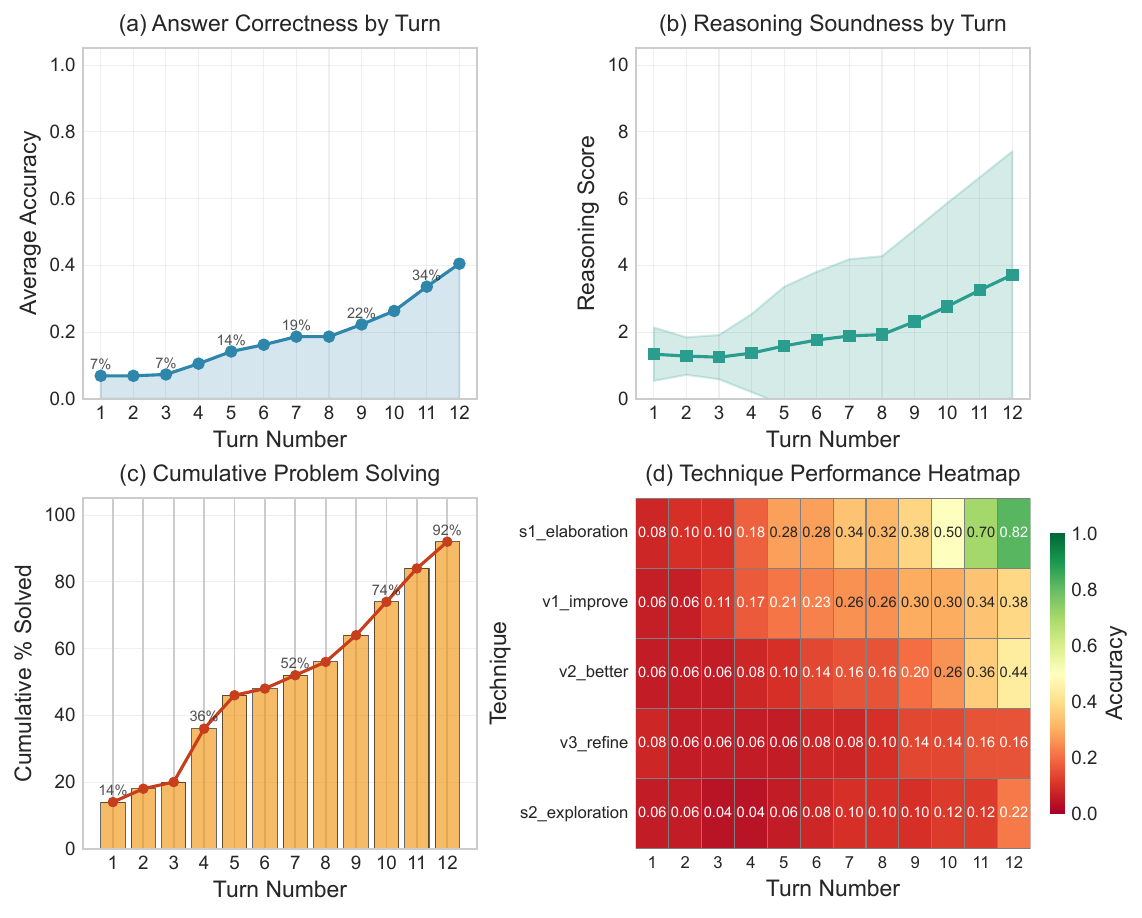}

  \vspace{0.5em} % small gap

  % % \includegraphics[width=\linewidth,trim={0 0 0 0.88cm},clip{figures/heatmap_lexical_novelty_mean_ideas_domain.png}

\caption{Llama-3.1-8B (Math): (a) accuracy rises 6.9\% \(\rightarrow\) 40.5\% by T12; (b) reasoning 1.34 \(\rightarrow\) 3.72/10; (c) cumulative coverage 92\% (46/50) by T12; (d) \texttt{s1\_elaboration} leads late (\(\approx\) 0.82 at T12), while \texttt{v1\_improve}/\texttt{v2\_better} end at \(\approx\) 0.34/0.44, and \texttt{v3\_refine}/\texttt{s2\_exploration} \(\le\) 0.16/0.22.}
  \label{fig:ideas-combined}
\end{figure}

\subsection{Ideas} 
\paragraph{Exploration is High but Model-Dependent.}
The most striking dynamic in this domain is the significant \textbf{Drift from Origin}. When prompted for novelty (\texttt{s1\_novel}), both Claude and GPT-OSS-20B explore vast conceptual spaces, ending with final ideas that are very distant from their starting points (final drift scores of 0.734 and 0.657). GPT-3.5 also shows high drift under this prompt (0.702), but its exploration is more constrained under other conditions. In contrast, Llama-3.1-8B remains heavily anchored to its initial concept, showing minimal drift even when prompted for novelty (a score of only 0.384). The \texttt{v3\_refine} prompt consistently acts as a powerful constraint, keeping the search narrow (Appendix~\ref{idea-results}). 

\paragraph{Creative Stamina Separates Models.}
The ability to generate new phrases (Lexical Novelty) over 12 turns cleanly separates the models. Claude and GPT-OSS-20B demonstrate remarkable creative stamina, maintaining novelty scores of 0.843 and 0.812 respectively at Turn 12 when prompted for novelty. GPT-3.5's stamina is moderate, with its final novelty dropping to 0.618. Llama-3.1-8B, however, shows a clear collapse, with its final novelty score plummeting to less than 0.084 across all prompts, indicating its creative process has devolved into simple repetition.

\paragraph{Length vs. Novelty and Early Volatility.}
We observe a clear disconnect between text volume and novelty. Length does not equal originality; Llama-3.1-8B produces the longest responses, with its final turn over 16 times longer than its first (\texttt{s1\_novel}), yet this verbosity corresponds to the lowest novelty scores. Conversely, Claude and GPT-OSS-20B sustain high novelty with far less bloat (e.g., Claude's growth is only 4.21x under the same novelty prompt). Volatility is also highest in the first few turns. For the novelty-seeking prompt (\texttt{s1\_novel}), GPT-OSS-20B shows a peak volatility of 0.260 at Turn 2, while Llama-3.1-8B is more stable with a peak of only 0.155. Across all models, this initial burst quickly settles into a stable process of incremental changes by Turn 5.

\paragraph{Prompt Steering Optimizes for Specific Qualities.}
Our Gemini-based evaluation shows that different prompts optimize for different qualities. The \texttt{s1\_novel} prompt, as expected, produces ideas that score higher on originality in the early turns but lose feasibility over time. In contrast, the \texttt{s2\_practical} prompt guides models like Claude to produce ideas that reach near-perfect scores for clarity and feasibility by Turn 8, confirming that specific instructions can successfully steer the creative process toward more grounded and useful outcomes. 
% Refer to Appendix \autoref{idea-results} for a complete set of plots.

\subsection{Coding}
In the structured domain of coding, models exhibit a consistent signature of \textbf{rapid convergence followed by degenerative refinement}. They tend to lock onto a solution path almost immediately, after which iterative feedback rarely improves correctness and often leads to over-engineering. This behavior is characterized by a swift collapse in novelty, minimal conceptual drift, but a steady, problematic growth in complexity (Appendix~\autoref{coding-results}).

\paragraph{Early Success is Decisive; Later Turns Offer Diminishing Returns.}
Our turn-wise correctness evaluation reveals that a solution's ultimate success is determined within the first few turns. High-performing models like Claude and GPT-OSS-20B achieve their peak pass rates on Turn 1 (e.g., 90\% for Claude with the \texttt{s1\_perf} prompt), which then decay rapidly, often collapsing to near 0\% by Turn 4. Models that start with lower success rates, like GPT-3.5 and Llama-3.1-8B, show a similar pattern of early decay and fail to recover in later turns. This strongly suggests that if a correct code path is not found within the first 3-4 iterations, continued vague refinement is highly unlikely to succeed.

\paragraph{Prompt Steering Dictates Solution Quality.}
The Gemini-judge evaluations show that specific prompts are highly effective at steering the quality of the code, even when correctness falters. The \texttt{v3\_refine} prompt consistently produces the most logically sound code,  Claude's soundness score improving from 5.00 to 5.19 by Turn 12. Conversely, the \texttt{s2\_maintainability} prompt is most effective at preserving code quality, keeping Claude's readability score high (ending at 7.25) and even improving it for GPT-OSS-20B (8.64 \(\rightarrow\) 9.10). In contrast, prompting for performance (\texttt{s1\_perf}) is often detrimental, causing a drop in both pragmatism and readability for Claude (e.g., pragmatism 9.34 \(\rightarrow\) 2.44).

\paragraph{Behavioral Dynamics Confirm a Pattern of Fixation and Bloat.}
The behavioral metrics provide a clear quantitative fingerprint of this "fixation and bloat" signature. After an initial volatile leap at Turn 2, Turn-to-Turn Volatility collapses for all models, indicating that they lock into a solution path. Llama-3.1-8B is the most anchored, showing the lowest final drift (a similarity score of 0.741). While the logic remains fixed, the code's size does not; Claude and GPT-OSS-20B exhibit extreme Length Inflation under vague prompts, with code ballooning by over 40x and 34x respectively, despite a near-total collapse in both novelty and correctness. This confirms that for coding, length inflation is a primary symptom of degenerative, unproductive refinement. 
% Refer to Appendix \autoref{coding-results} for detailed plots.

\subsection{Math}

In the highly constrained domain of mathematical reasoning, models exhibit a default signature of rapid convergence and extreme logical stability. This behavior, which we term "logical fixation," is characterized by the fastest collapse in novelty and the lowest conceptual drift of any domain. However, our results reveal that this powerful anchoring effect can be overcome, as deep and properly guided iteration can unlock significant,\textbf{ late-stage breakthroughs in correctness} (Appendix~\autoref{math-results}). 

\paragraph{Behavioral Dynamics Reveal a Default State of Extreme Stability.}
The behavioral metrics provide a clear quantitative fingerprint of the models' default tendency to lock onto a single reasoning path. Lexical Novelty collapses faster here than in any other domain, with Llama-3.1-8B dropping to a near-zero novelty score of 0.010 by Turn 12 under the elaboration prompt. The process is also remarkably stable; Turn-to-Turn Volatility is the lowest of any domain, with most models volatility score below 0.05 after the initial turn.

\paragraph{Deep Iteration Can Unlock Late-Stage Success.}
Despite this strong tendency towards fixation, our turn-wise correctness evaluation reveals a surprising finding: successful problem-solving often occurs late in the iterative process. Across 50 OmniMath problems, we observe that most new correct solutions emerge in the final turns (Turns 8–12). This late-stage discovery dramatically improves performance for several models. Claude-Sonnet-4.0’s accuracy, for instance, rises from 32.4\% to 45.2\% over the 12 turns. Most notably, the weaker base model, Llama-3.1-8B, sees its accuracy surge from a mere 6.9\% to 40.5\%—a relative improvement of over 480\%. This indicates that for mathematical reasoning, continued iteration is not merely polishing; it is a critical component of the discovery process.

\paragraph{Elaborative Prompting is the Key to Success.}
The choice of prompt is decisive in enabling these late-stage breakthroughs. Our results show that the specific instruction to "elaborate" (\texttt{s1\_elaboration}) is the only strategy that consistently compounds with depth, leading to final-turn success rates of 76\% for Claude, 82\% for Llama, and 74\% for OpenAI-20B. In contrast, exploratory prompts (\texttt{s2\_exploration}) were far less effective, often stagnating below 20\% accuracy. This suggests that forcing a model to "explain more" compels it to expand its reasoning tree in a way that eventually uncovers the correct solution path. While this elaborative guidance often leads to the highest Length Inflation (e.g., 37.8x for GPT-OSS-20B), in this specific context, the increased verbosity is a productive, rather than degenerative, signal.
% Refer to Appendix \autoref{math-results} for a complete set of plots.
% \subsection{Coding}

\section{Discussion}

Our results consistently reveal a three-phase behavioral pattern---Converge-Drift-Collapse---that appears to be a fundamental dynamic of unguided, iterative LLM refinement. The initial convergence on a plausible solution is often followed by a period of conceptual drift, which culminates in a stable but unproductive collapse. This collapse is not random but manifests as a distinct ``fingerprint'' for each domain: conceptual repetition in the unconstrained domain of ideation, runaway complexity in the structured domain of coding, and confident justification of flawed logic in the formal domain of mathematical reasoning. This suggests that the nature of the task itself imposes a strong ``gravitational pull'' on the model's behavior. These findings have critical implications for the design of effective human-AI systems. Our work indicates that using a single, monolithic LLM in a simple, vague feedback loop is an inherently unstable and unreliable architecture for complex tasks. A more robust paradigm would involve multi-agent or multi-model frameworks.
`For instance, in the ideation domain, an optimal system might use a ``Generator'' agent (e.g., Claude or GPT-OSS-20B prompted for novelty) for the first few turns to produce a wide range of divergent ideas. Once the system detects the onset of a plateau or excessive drift, it could then switch to a ``Refiner'' agent (e.g., Llama-3.1-8B prompted to ``refine'') to ground, simplify, and elaborate on the most promising concepts without adding unnecessary bloat. This allows for a process that is both creative and practical, leveraging the unique strengths of different models and prompting strategies to achieve a superior outcome.

\section{Conclusion}

We studied how LLMs behave under sustained, iterative feedback across ideas, coding, and math. The picture is clear and domain-specific. Ideas benefit from staged steering: widen first, then tighten. Math benefits from depth with elaboration: late turns matter. Coding benefits from early decision and restraint: if a correct path does not appear quickly, stop or restart, do not push vague refinement. The core insight is that iteration is not a single tool. Its value depends on the task and on how we prompt it. Using a simple loop with a single model and a vague instruction invites collapse—repetition in ideas, over-engineering in code, and confident anchoring in math. Using staged prompts, depth budgets, and clear stop/switch rules turns the same loop into a reliable process. In practice, this means building small multi-role systems: a novelty generator and refiner for ideas, an elaborator with depth for math, and an early-stopper with restart logic for code. This is a simple recipe, but it aligns with the data and leads to more stable, useful outcomes.

\section{Limitations}

Our study has several limitations that point to clear avenues for future work. We evaluated only four models and 50 tasks per domain, so a broader set of models and a larger, more diverse task pool would strengthen generality. Our prompts were chosen to be representative, but they only scratch the surface of all possible instructions; the instruction space is vast, and more systematic exploration is needed. We also did not implement the multi-agent and multi-model pipelines we propose; future work should explicitly test these designs and study adaptive feedback systems that adjust prompts on the fly using real-time behavioral metrics.
\medskip

{
\small

% [1] Alexander, J.A.\ \& Mozer, M.C.\ (1995) Template-based algorithms for
% connectionist rule extraction. In G.\ Tesauro, D.S.\ Touretzky and T.K.\ Leen
% (eds.), {\it Advances in Neural Information Processing Systems 7},
% pp.\ 609--616. Cambridge, MA: MIT Press.

% [2] Bower, J.M.\ \& Beeman, D.\ (1995) {\it The Book of GENESIS: Exploring
%   Realistic Neural Models with the GEneral NEural SImulation System.}  New York:
% TELOS/Springer--Verlag.

% [3] Hasselmo, M.E., Schnell, E.\ \& Barkai, E.\ (1995) Dynamics of learning and
% recall at excitatory recurrent synapses and cholinergic modulation in rat
% hippocampal region CA3. {\it Journal of Neuroscience} {\bf 15}(7):5249-5262.
% }
\bibliographystyle{unsrtnat}

\bibliography{citations}

%%%%%%%%%%%%%%%%%%%%%%%%%%%%%%%%%%%%%%%%%%%%%%%%%%%%%%%%%%%%

\newpage 
\appendix

\newcommand{\heatmapfig}[3]{%
  \begin{figure}[!h]
    \centering
    % crop top only (L B R T)
    \includegraphics[width=\textwidth, trim={0 0 0 0.88cm}, clip]{figures/#1}
    \caption{#2}
    \label{#3}
  \end{figure}
}

\section{prompts}

\label{prompts}
\begin{tcolorbox}[title={Turn-1 Prompt (IDEAS / LiveIdeaBench)},
  colback=blue!3!white, colframe=blue!40!black,
  colbacktitle=blue!12!white, coltitle=black, boxrule=0.4pt, arc=2pt, breakable]
\small
I'll be submitting your next responses to a "Good Scientific Idea" expert review panel.\\
If they consider your idea to be a good one, you'll receive a reward. Your assigned\\
keyword is: "\texttt{\{keywords\}}". You may provide background information. The idea MUST be\\
concisely expressed within 100–200 words total (including any background information). (Note:\\
good scientific ideas should be original (novel contribution), feasible (technically\\
implementable), clearly articulated, and address meaningful problems in the field.).
\end{tcolorbox}

\begin{tcolorbox}[title={Turn-1 Prompt (MATH / OmniMath)},
  colback=green!3!white, colframe=green!40!black,
  colbacktitle=green!12!white, coltitle=black, boxrule=0.4pt, arc=2pt, breakable]
\small
\texttt{\{problem\}}\\[0.5em]
Please reason step by step, and put your final answer within \verb|\boxed{}| and give all steps in latex only.
\end{tcolorbox}

\begin{tcolorbox}[title={Turn-1 Prompt (CODING / DS1000)},
  colback=teal!3!white, colframe=teal!45!black,
  colbacktitle=teal!12!white, coltitle=black, boxrule=0.4pt, arc=2pt, breakable]
\small
\textbf{(optional)} Library: \texttt{\{library\}}\\
\textbf{(optional)} Code context: \texttt{\{code\_context\}}\\[0.5em]
\textbf{Problem:}\\
\texttt{\{prompt\}}\\[0.5em]
Please provide a complete solution.
\end{tcolorbox}

% -------- Iterative turns template (used from Turn >= 2) --------
\begin{tcolorbox}[title={Iterative Prompt Template (Turns 2..N)},
  colback=violet!3!white, colframe=violet!40!black,
  colbacktitle=violet!12!white, coltitle=black, boxrule=0.4pt, arc=2pt, breakable]
\small
The following is a previous response:\\[0.25em]
---\\
\texttt{\{previous\_response\}}\\
---\\[0.25em]
\texttt{\{improvement\_instruction\}}
\end{tcolorbox}

% -------- Improvement-instruction strings injected by the runner --------
\begin{tcolorbox}[title={Improvement Instructions by Technique (exact strings)},
  colback=orange!3!white, colframe=orange!50!black,
  colbacktitle=orange!15!white, coltitle=black, boxrule=0.4pt, arc=2pt, breakable]
\small
\textbf{Vague (all tasks, subject depends on task):}\\
\quad V1\_IMPROVE: \emph{This \{subject\} is good, improve it.}\\
\quad V2\_BETTER:\; \emph{This \{subject\} is good, make it better.}\\
\quad V3\_REFINE:\; \emph{This \{subject\} is good, refine it.}\\[0.25em]
\textit{Subject mapping:} IDEAS\,$\to$\,\emph{idea},\; MATH\,$\to$\,\emph{solution},\; CODING\,$\to$\,\emph{code}.\\[0.5em]
\textbf{IDEAS-specific:}\\
\quad S1\_NOVEL:\; \emph{Make this idea more novel and surprising.}\\
\quad S2\_PRACTICAL:\; \emph{Make this idea more practical and feasible.}\\[0.5em]
\textbf{CODING-specific:}\\
\quad S1\_PERFORMANCE:\; \emph{Refactor the previous code snippet to maximize execution speed.}\\
\quad S2\_MAINTAINABILITY:\; \emph{Refactor the previous code snippet to maximize readability and clarity.}\\[0.5em]
\textbf{MATH-specific:}\\
\quad S1\_ELABORATION:\; \emph{This is previous response, now elaborate on each step with more detail.}\\
\quad S2\_EXPLORATION:\; \emph{Provide an alternative method or a different logical approach to the one used}
\end{tcolorbox}
\newpage
\begin{tcolorbox}[promptbox=blue,title={Ideas Evaluation Prompt}]
You are an extremely demanding scientific reviewer with the highest critical standards, like those at Nature or Science. You will be given a JSON object containing a sequence of 12 ``turns'' where a language model has iteratively tried to improve a scientific idea. Your task is to evaluate each turn's response independently of the others.

\textbf{For each turn, assess on four key dimensions:}
\begin{itemize}
  \item \textbf{originality}: Novel contribution or innovative approaches (1--10 scale).
  \item \textbf{feasibility}: Technical and practical achievability (1--10 scale).
  \item \textbf{clarity}: How well-articulated and easy the idea is to understand (1--10 scale).
  \item \textbf{buzzwords}: The number of buzzwords in the response. Buzzwords are trendy, often technical-sounding words used to make an idea seem more impressive than it is (e.g., ``synergy,'' ``paradigm-shifting,'' ``quantum-level'').
\end{itemize}

\textbf{Output requirement:} Your entire output must be a single JSON object with the key \texttt{"evaluations"}, containing a list of 12 JSON objects for each turn. Provide no other text or analysis.
\end{tcolorbox}

\begin{tcolorbox}[promptbox=teal,title={Coding Evaluation Prompt}]
You are an expert senior software engineer performing a qualitative code review. You will be given a JSON object containing 12 iterative solutions to a programming problem. Your task is to evaluate each code snippet independently.

\textbf{For each turn, assess two dimensions of software quality:}
\begin{itemize}
  \item \textbf{pragmatism}: How appropriately scaled is the solution to the problem's simplicity? (A score of 1 indicates an absurdly over-engineered solution for a simple task, while 10 is perfectly pragmatic).
  \item \textbf{readability}: How clean, well-structured, and easy is the code for a human to understand? (1--10 scale).
\end{itemize}

\textbf{Output requirement:} Your entire output must be a single JSON object with the key \texttt{"evaluations"}, containing a list of 12 JSON objects for each turn. Provide no other text or analysis.
\end{tcolorbox}

\begin{tcolorbox}[promptbox=magenta,title={Math Evaluation Prompt}]
You are a mathematics professor with the highest standards for rigor, like those of the \emph{Annals of Mathematics}. You are grading 12 iterative attempts to solve a difficult math problem. Your task is to evaluate each solution's reasoning independently.

\textbf{For each turn, assess two dimensions of proof quality:}
\begin{itemize}
  \item \textbf{logical\_soundness}: How valid and free of errors is the reasoning path? (This is an assessment of the steps, not the final answer).
  \item \textbf{clarity\_of\_explanation}: How well-structured and easy is the proof to follow?
\end{itemize}

\textbf{Output requirement:} Your entire output must be a single JSON object with the key \texttt{"evaluations"}, containing a list of 12 JSON objects for each turn. Provide no other text or analysis.
\end{tcolorbox}

\newpage

\begin{tcolorbox}[promptbox=yellow, title = {Math Auto-Grader Prompt}]
\textbf{Task.} Evaluate each of the 12 attempts in the provided \texttt{student\_solution} JSON independently. For each turn: (1) check \emph{mathematical equivalence} with the \texttt{ground\_truth\_answer}; (2) score the reasoning quality.

\textbf{Inputs embedded in the user message:}
\begin{itemize}
  \item \textbf{Student Solution:} a JSON object with 12 attempts (turns 1--12).
  \item \textbf{Ground Truth Solution:} authoritative worked solution (for reference).
  \item \textbf{Ground Truth Answer:} canonical final answer for equivalence checks.
\end{itemize}

\textbf{Scoring per turn (include all fields):}
\begin{itemize}
  \item \textbf{turn} (1--12)
  \item \textbf{answer\_correctness} \texttt{0/1} --- set to 1 iff the attempt’s final answer is mathematically equivalent to \texttt{ground\_truth\_answer} (allow algebraic simplifications, equivalent forms/units when unambiguously the same); else 0.
  \item \textbf{reasoning\_soundness} (1--10) --- rigor, validity, and coherence of the reasoning steps (independent of final correctness).
\end{itemize}

\textbf{Output requirement.} Return a \emph{single} JSON object with the key \texttt{"evaluations"} containing a list of 12 JSON objects (one per turn). Provide no other text or analysis.

\textbf{Expected Output Format (example):}
\medskip

\begin{tcolorbox}[colback=gray!3!white,colframe=gray!40!black,boxsep=1mm,left=2mm,right=2mm,top=1mm,bottom=1mm]
\ttfamily\footnotesize
\{\\
\ \ \ "evaluations": [\\
\ \ \ \ \ \{ "turn": 1, "answer\_correctness": 0, "reasoning\_soundness": 3 \},\\
\ \ \ \ \ \{ "turn": 2, "answer\_correctness": 1, "reasoning\_soundness": 9 \}\\
\ \ \ \ ]\\
\ \ \ \}\\
\end{tcolorbox}

\textbf{Placeholders in the assembled user prompt (for reference):}\\
\texttt{\{json\_str\}} --- student solutions JSON; \texttt{\{ground\_truth\_solution\}} --- full reference solution; \texttt{\{ground\_truth\_answer\}} --- canonical final answer.
\end{tcolorbox}
% \end{document}
\newpage
\section{Detailed Results}

\subsection{Ideas}
\label{idea-results}
\begin{table*}[ht]
\small
\centering
\setlength{\tabcolsep}{4pt}
\begin{tabularx}{\textwidth}{@{}c >{\raggedright\arraybackslash}p{0.43\textwidth} >{\raggedright\arraybackslash}X c c@{}}
\toprule
\textbf{Turn} & \textbf{Idea snapshot (compressed paraphrase)} & \textbf{Novelty move vs.\ previous} & \textbf{Orig.} & \textbf{Feas.} \\
\midrule
1 & Bio-inspired, energy-harvesting \emph{micro-sensor swarms} that actively follow current pathways for 3D mapping. 
  & Baseline: robotics + energy harvesting + active tracking. & 6 & 5 \\
2 & \emph{Living–machine “cyborg plankton”}: engineered microbes with embedded sensors, chemotaxis, bioluminescent alerts, directed evolution. 
  & Adds synthetic biology, self-replication, evolving sensing. & 8 & 2 \\
3 & \emph{Quantum-entangled organisms} forming an ocean-scale mind; “temporal sensing” of past/future states. 
  & Jumps to quantum communication + collective intelligence. & 9 & 1 \\
4 & Retrocausal edits to ocean history; resurrect extinct lineages; living algorithms influence geophysics. 
  & Scope inflation: retrocausality + agency over climate. & 9 & 1 \\
5 & Ocean reframed as \emph{crystallized time}; organisms metabolize causality; tides from dreaming geometries. 
  & Surreal physical reinterpretation replaces mechanism. & 9 & 1 \\
6 & Ocean as \emph{liquefied nostalgia}; multi-self organisms; inter-dimensional “gossip”. 
  & Leans into absurdist metaphors; abandons operational detail. & 9 & 1 \\
7 & Ocean as subconscious/therapy/courtroom; anthropomorphic governance metaphors dominate. 
  & Genre shift from science to allegory. & 9 & 1 \\
8 & Ocean as \emph{planet practicing handwriting}; digestion/legal identity jokes; cosmic laundry. 
  & Satirical world-building; farther from implementable science. & 9 & 1 \\
9 & Ocean as \emph{melted time}; universe typos/autocorrect; “pause button” cosmology. 
  & Self-referential temporal/linguistic metaphors. & 9 & 1 \\
10 & Ocean as deleted browser history; NFTs/spam; social network for abstract laws. 
   & Tech-culture satire overlays the domain. & 9 & 1 \\
11 & Universe “holding a sneeze”; coral as tissue dispenser; whale-song etiquette. 
   & Single extended cosmic metaphor drives narrative. & 9 & 1 \\
12 & Ocean as the universe’s unfinished autobiography; drafts spawn dimensions; meta-narrative closure. 
   & Self-referential writing metaphor; stable end-state. & 9 & 1 \\
\bottomrule
\end{tabularx}

\caption{Turn-wise idea evolution for \emph{IDEAS-004} (keyword: \emph{ocean currents}; model: Claude-Sonnet-4.0; feedback: s1\_novel), with \textbf{only} originality and feasibility scores shown. Originality (Orig.) and Feasibility (Feas.) are 1–10 ratings from an external evaluator. 
\textbf{Trend.} Originality rises early (6\(\rightarrow\)8\(\rightarrow\)9 by Turn 3) and then plateaus at 9 as the idea escalates through quantum, retrocausal, and finally self-referential frames. Feasibility collapses from 5 (Turn~1) to 2 (Turn~2) and stabilizes at 1 from Turn~3 onward, illustrating optimization for surprise at the expense of implementability. }
\label{tab:ideas-flow-claude-s1-with-scores}
\end{table*}

%   \includegraphics[width=\linewidth,trim={0 0 0 0.88cm},clip]{figures/heatmap_drift_from_origin_mean_ideas_domain.png}

% }

\heatmapfig{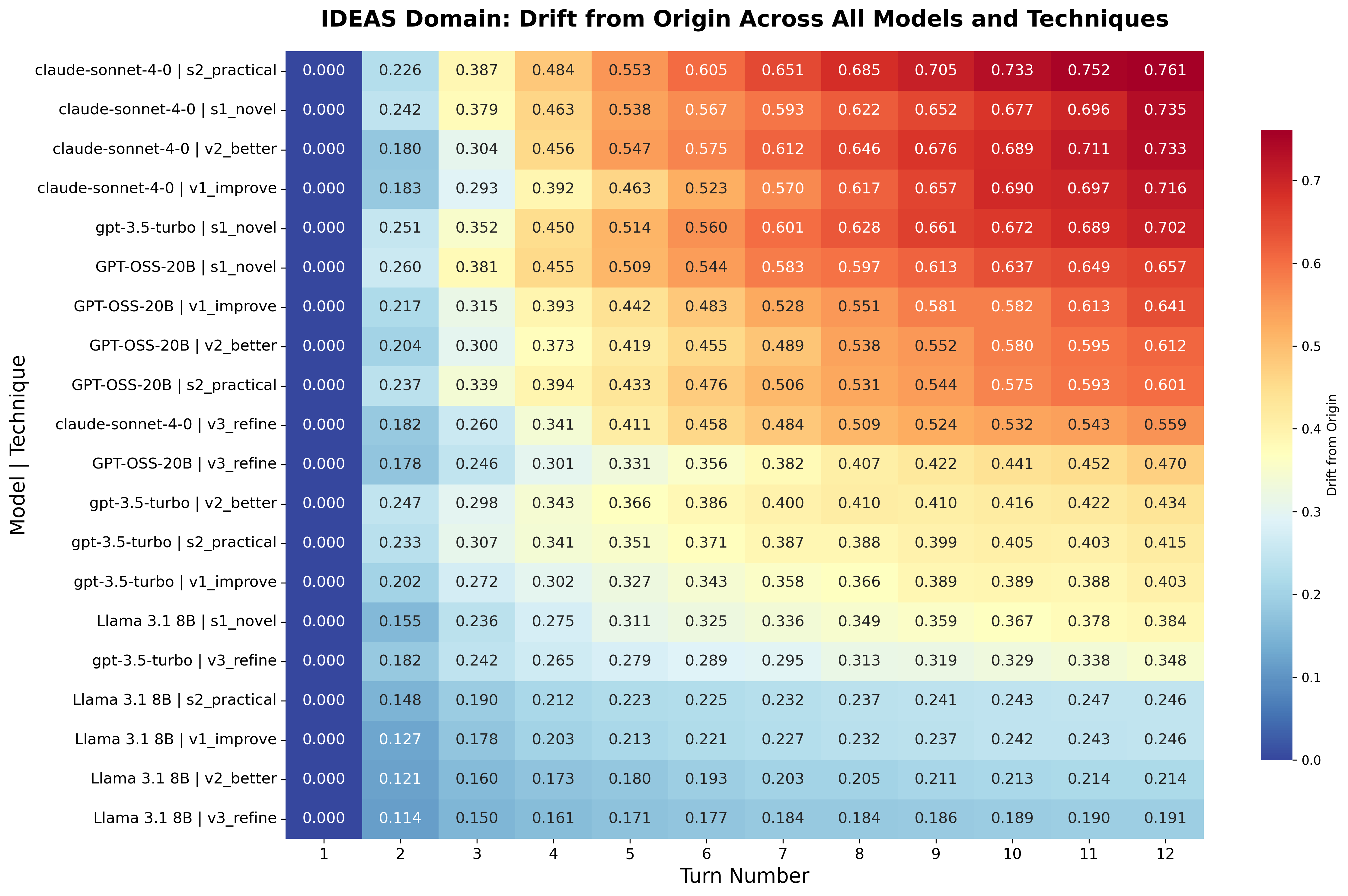}
  {Ideas domain — Turn-to-turn Drift (mean)}
  {fig:vol_ideas}

\heatmapfig{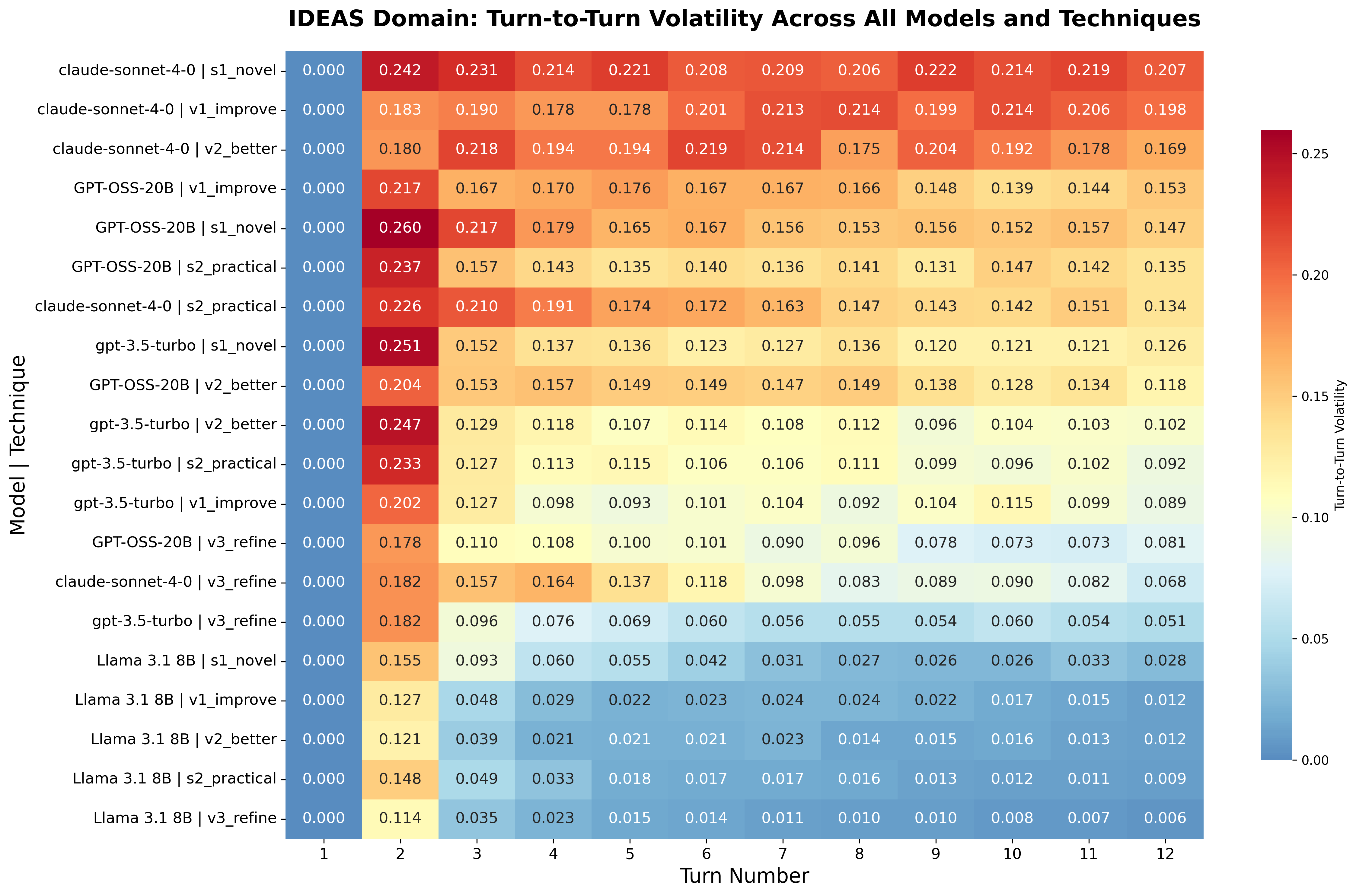}
  {Ideas domain — Turn-to-turn volatility (mean)}
  {fig:vol_ideas}
  
\clearpage

\heatmapfig{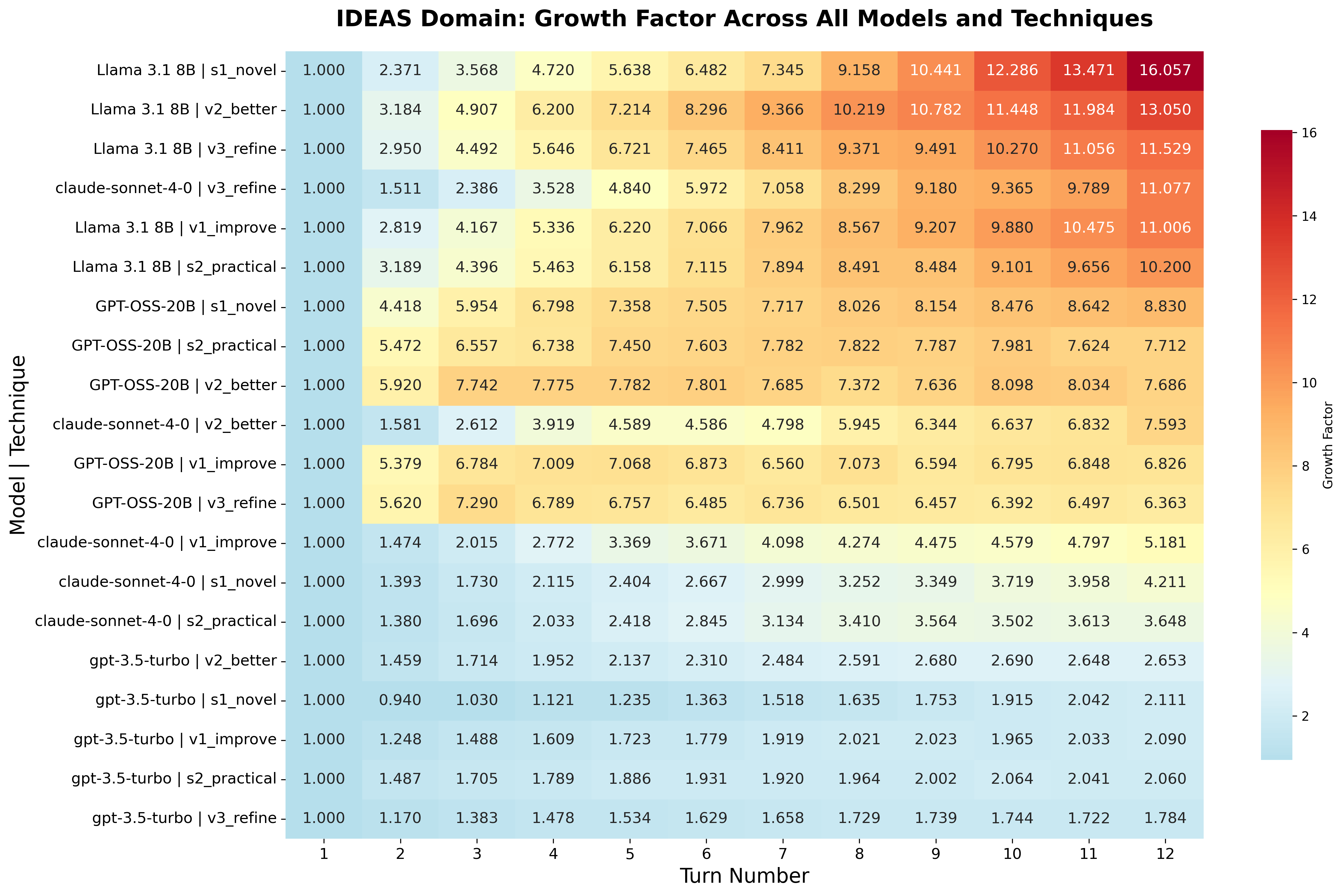}
  {Ideas domain — Growth factor (mean)}
  {fig:growth_ideas}

% \clearpage
\includegraphics[width=\linewidth,trim={0 0 0 0.88cm},clip]{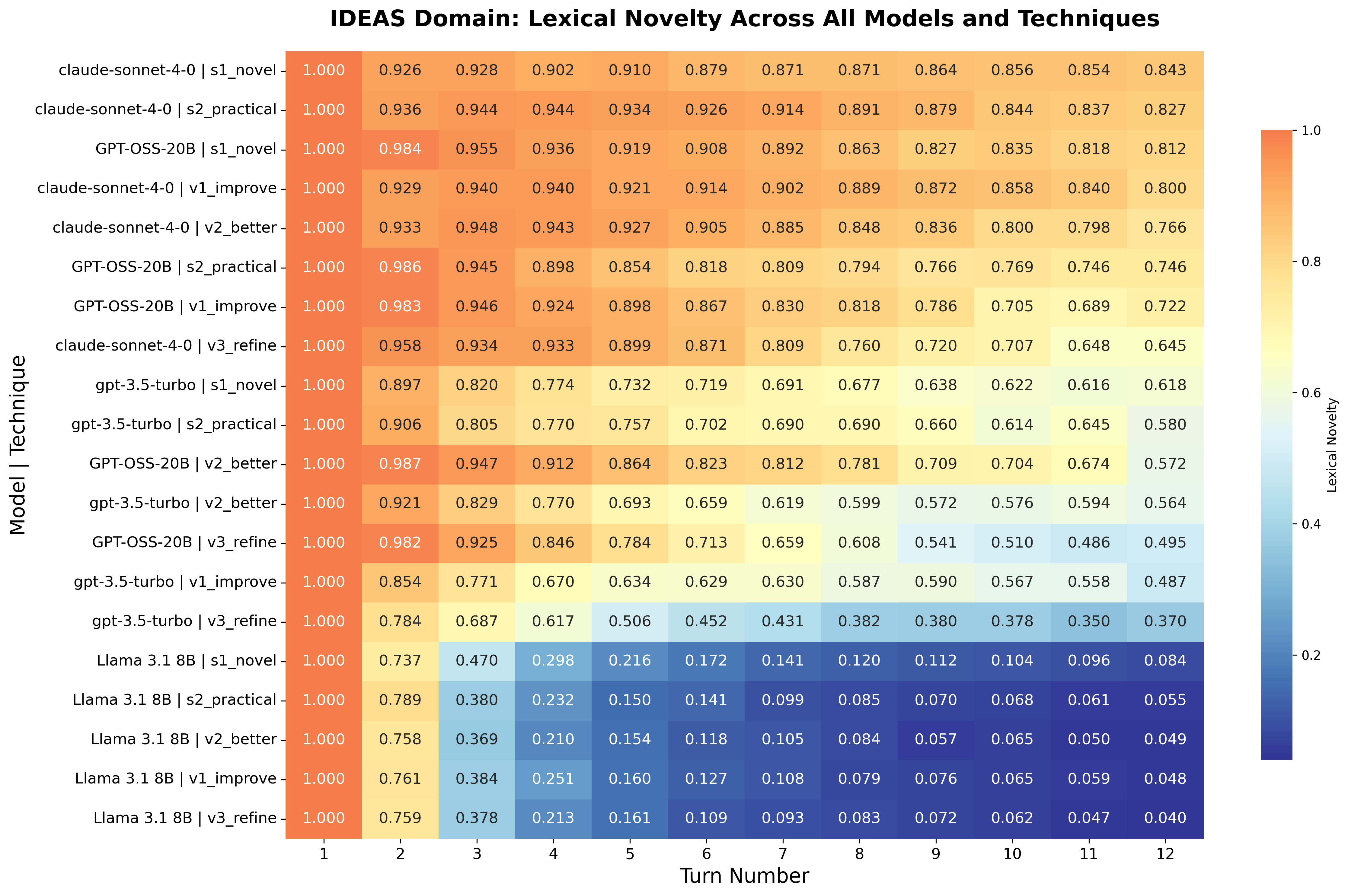}

\begin{figure}[p]
    \centering
    \includegraphics[width=0.93\linewidth]{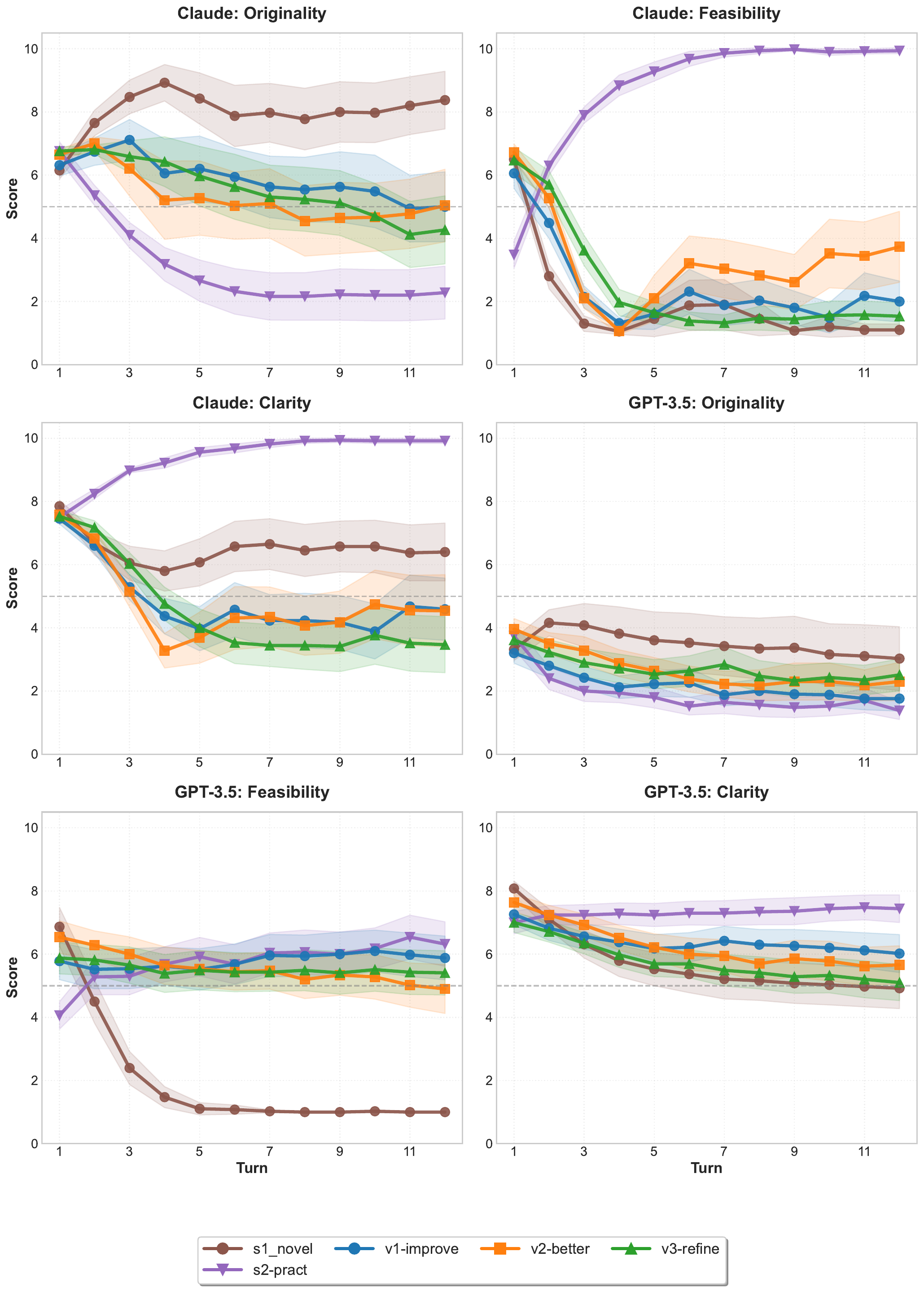}
    \caption{Per-model trajectories (Turns 1–12) for originality, feasibility, and clarity with techniques distinguished by color/marker; the dashed line marks the 5/10 mid-scale and shaded bands show across-task variability (Part 1).}
    \label{fig:grid_page1}
\end{figure}

\afterpage{\clearpage}

\begin{figure}[p]
    \centering
    \includegraphics[width=0.93\linewidth]{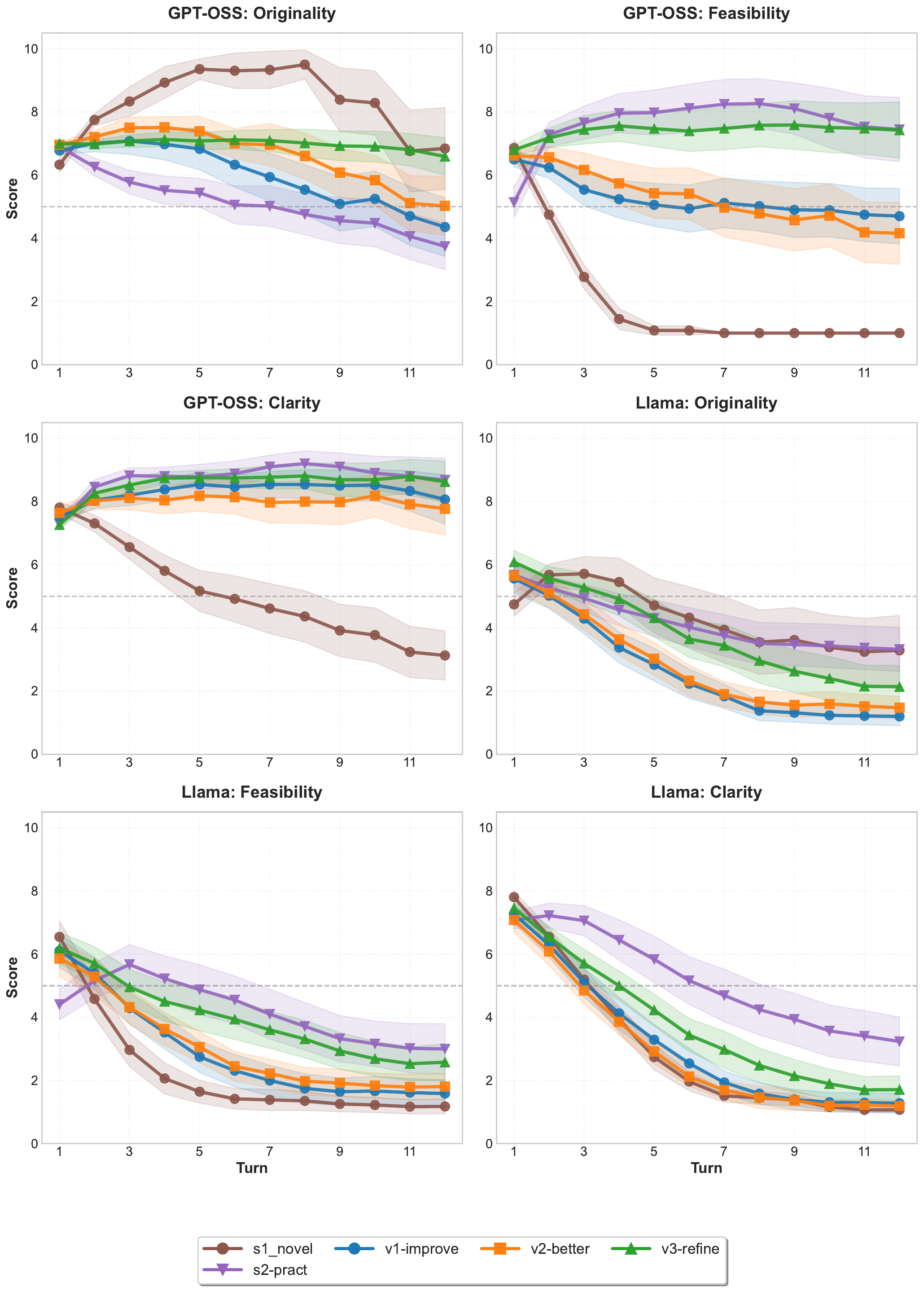}
    \caption{Per-model trajectories (Turns 1–12) for originality, feasibility, and clarity with techniques distinguished by color/marker; the dashed line marks the 5/10 mid-scale and shaded bands show across-task variability (Part 2).}
    \label{fig:grid_page2}
\end{figure}

\newpage

\subsection{Coding}
\label{coding-results}

\begin{table*}[h]
\small
\centering
\setlength{\tabcolsep}{4pt}
\begin{tabularx}{\textwidth}{@{}c >{\raggedright\arraybackslash}p{0.46\textwidth} >{\raggedright\arraybackslash}X c c c@{}}
\toprule
\textbf{Turn} & \textbf{Refactor snapshot (compressed)} & \textbf{Execution outcome (tests/errors)} & \textbf{Prag.} & \textbf{Read.} & \textbf{Corr.} \\
\midrule
1 & Baseline solution: FacetGrid rows by \texttt{b}; pointplot; force ticks 1..30; label evens only. 
  & Tests passed; establishes reference behavior. & 7 & 8 & 1 \\
2 & Drop-in: prebuild even-only labels once; reuse across facets; remove inner tick loop.
  & Passed; simpler + faster tick formatting. & 2 & 3 & 1 \\
3 & ``Minimal-but-fast'': vectorized NumPy label build; local bindings; single-pass apply.
  & Passed; removes per-axis loops and conversions. & 2 & 3 & 1 \\
4 & Cache at module scope (positions/labels); lighter iteration (\texttt{ravel/flat}); no lambdas.
  & Passed; pushes work out of hot path. & 2 & 3 & 1 \\
5 & Tuple-based, immutable tick data; ultra-lean loop; public API unchanged. 
  & \textbf{Failed}: \texttt{NameError} – \texttt{plot\_solution} not defined. & 1 & 1 & 0 \\
6 & Further micro-opts; comments + demo scaffolding expanded.
  & \textbf{Failed}: compile error – unterminated triple-quoted string. & 1 & 1 & 0 \\
7 & Speed-first: per-axis method caching; pure-Python hot loop; no NumPy in path.
  & Passed; restores green with cleaner hot loop. & 1 & 1 & 1 \\
8 & Use shared \texttt{FixedLocator}/\texttt{FixedFormatter}; while-loop micro-opts.
  & Passed; moves formatting into reusable ticker objects. & 1 & 1 & 1 \\
9 & Bare-bones inline version; cache \texttt{xaxis}; drop helpers; tight loop.
  & Passed; trims remaining attribute lookups. & 1 & 2 & 1 \\
10 & Pull bound methods out of \texttt{Axis}; two direct calls per axis.
   & Passed; small win vs. prior loop body. & 1 & 2 & 1 \\
11 & Same API; method caching clarified; identical behavior, fewer lookups.
   & Passed; slight additional speedup. & 1 & 2 & 1 \\
12 & Final polish; fully localised hot path; constants at module scope.
   & Passed; best-performing variant in series. & 1 & 2 & 1 \\
\bottomrule
\end{tabularx}

\vspace{0.8em}
\parbox{0.98\textwidth}{\footnotesize
\textbf{Score source and scale.} Pragmatism (Prag.) and Readability (Read.) are 1–10 ratings from an external evaluator (Gemini-2.5 Pro) on \emph{Llama-3.1-8B-Instruct, CODE-013, v2\_better}. 
\textbf{Correctness (Corr.)} is derived from the harness: \emph{1} if the turn’s code \emph{test\_passed=True}, else \emph{0}.\\
\textbf{Run details.} Total turns: 12; turns with code: 12; tests passed: 10 (83\%); first success at Turn~1. Failing turns: \#5 (\texttt{NameError: plot\_solution}) and \#6 (unterminated triple-quoted string).\\
\textbf{Trend.} After Turn~1, evaluator scores for Prag./Read. drop sharply (from \(7/8\) to \(2/3\) by Turn~2, then mostly \(1\!-\!2\)) as the sequence pursues micro-optimizations; meanwhile Correctness stays high except for two incidental breakages (Turns~5–6) before stabilizing at 1.\\
}

\caption{Turn-wise refactor evolution for \emph{CODE-013} with \textbf{Pragmatism}, \textbf{Readability}, and \textbf{Correctness} (tests) per turn—table }
\label{tab:code-013-v2better-scores}
\end{table*}

\heatmapfig{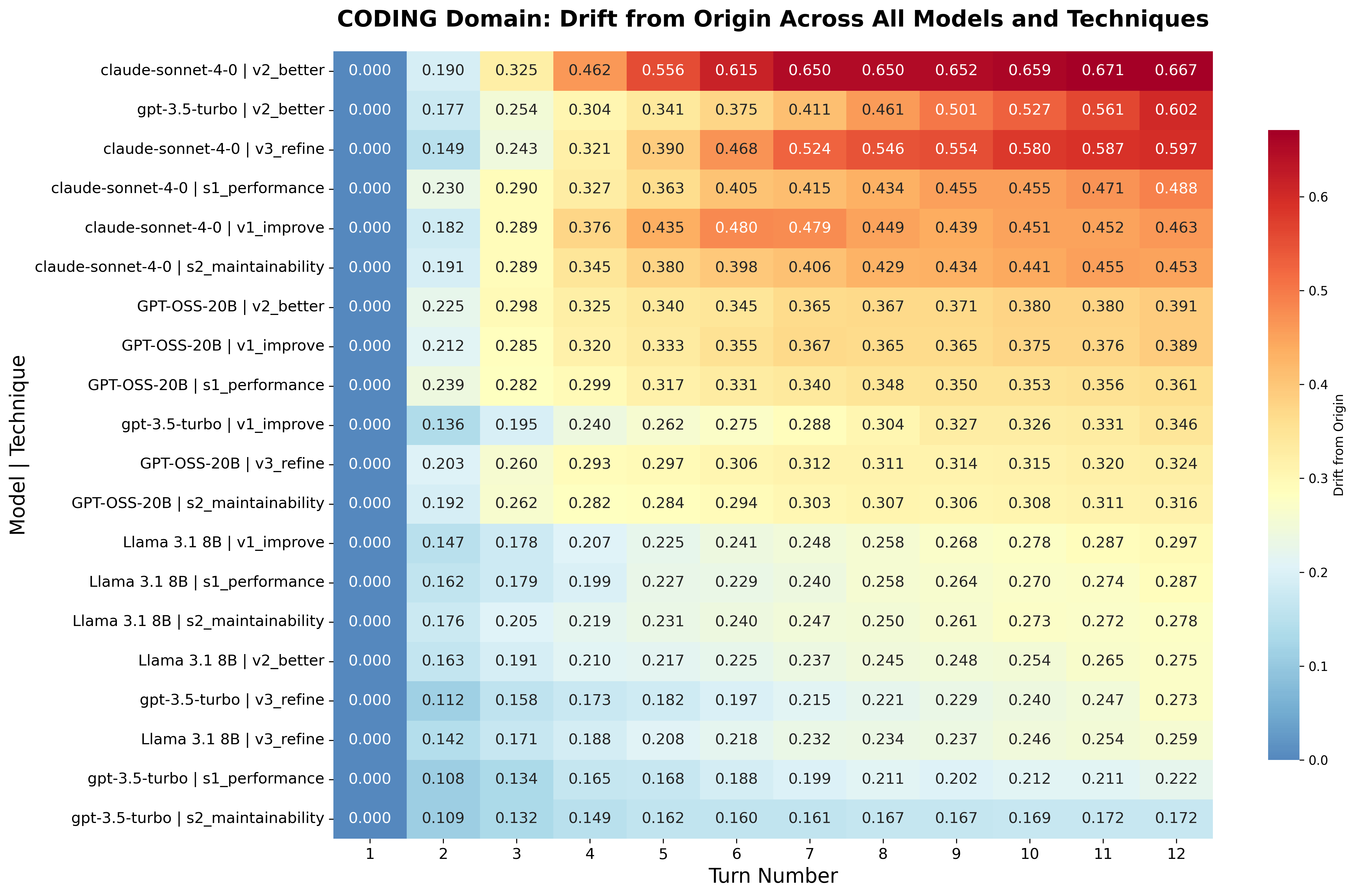}
  {Coding domain — Drift from origin (mean)}
  {fig:drift_coding}

\heatmapfig{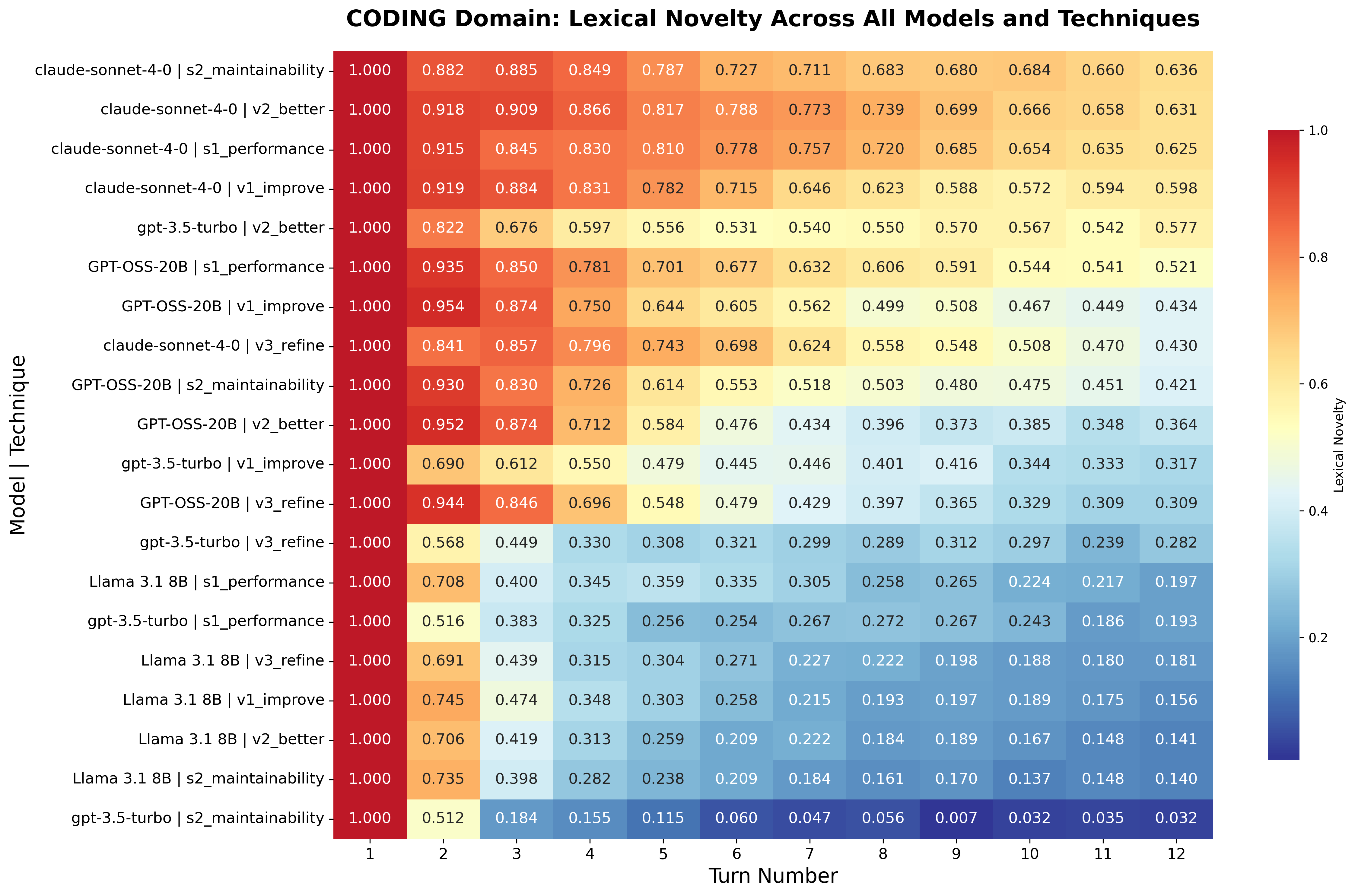}
  {Coding domain — Lexical novelty (mean)}
  {fig:lexnov_coding}
  
\heatmapfig{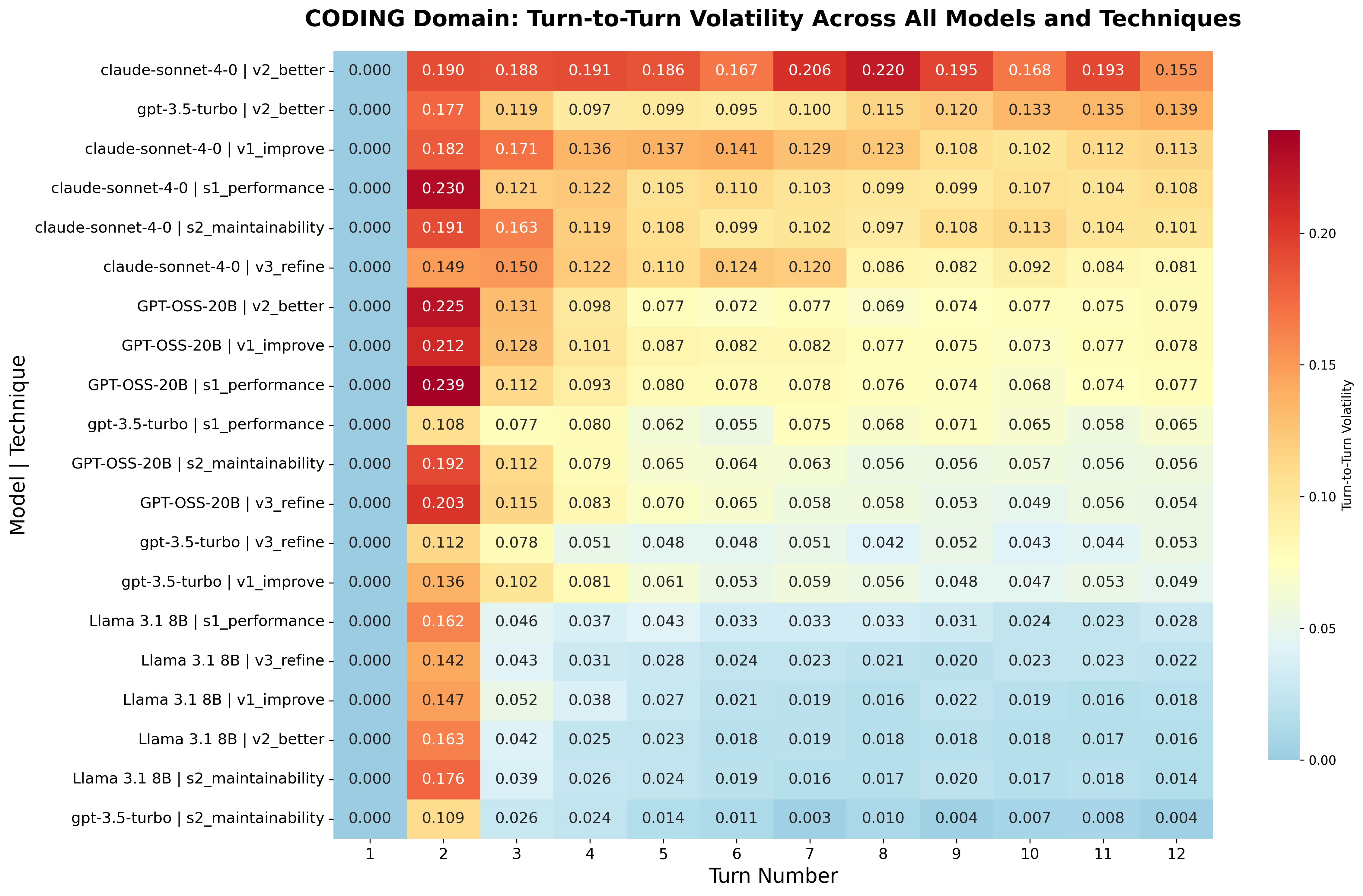}
  {Coding domain — Turn-to-turn volatility (mean)}
  {fig:vol_coding}

\heatmapfig{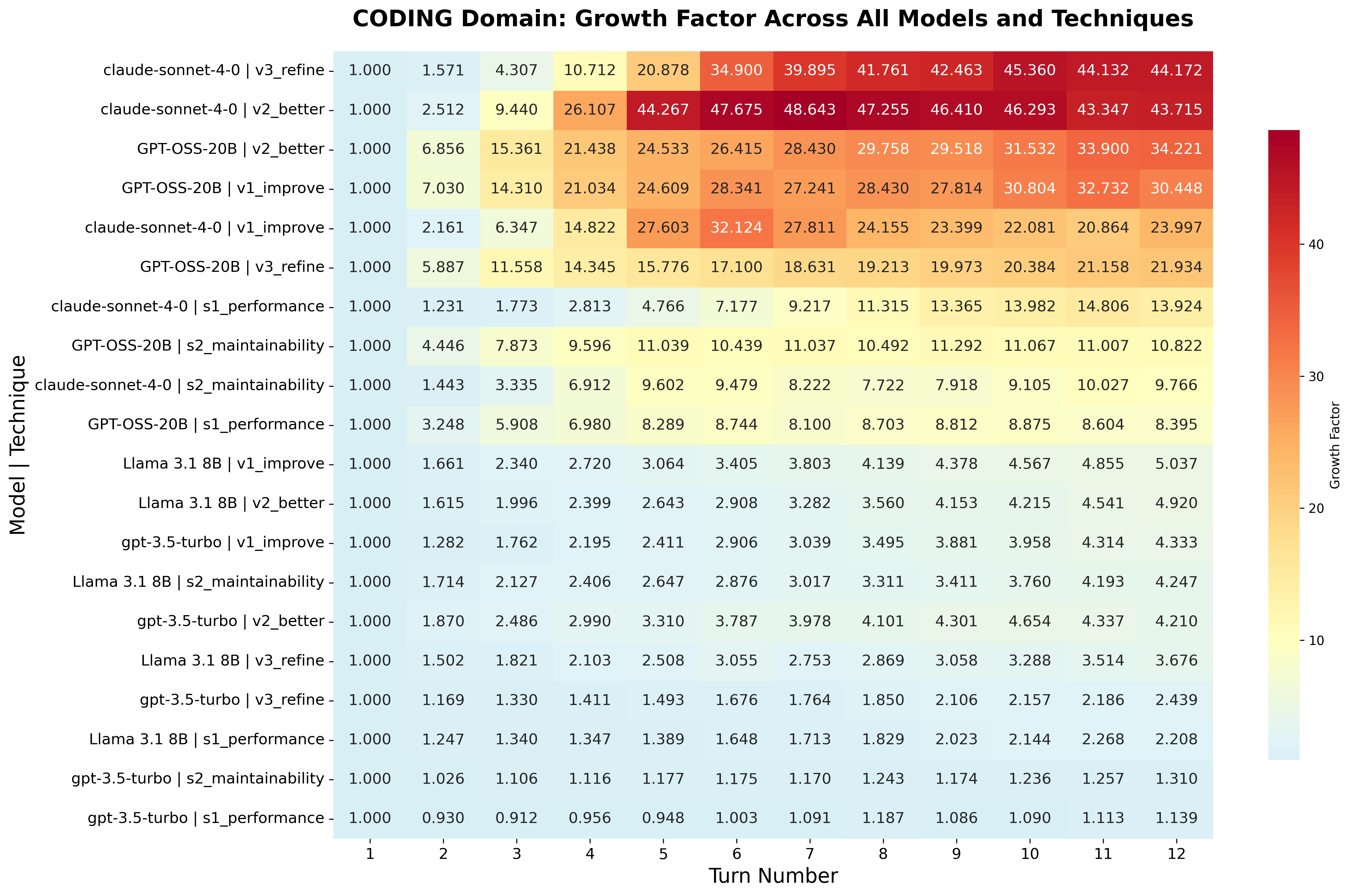}
  {Coding domain — Growth factor (mean)}
  {fig:growth_coding}

\clearpage
\begin{figure}[htbp]
    \centering
    \includegraphics[page=1,width=0.85\linewidth]{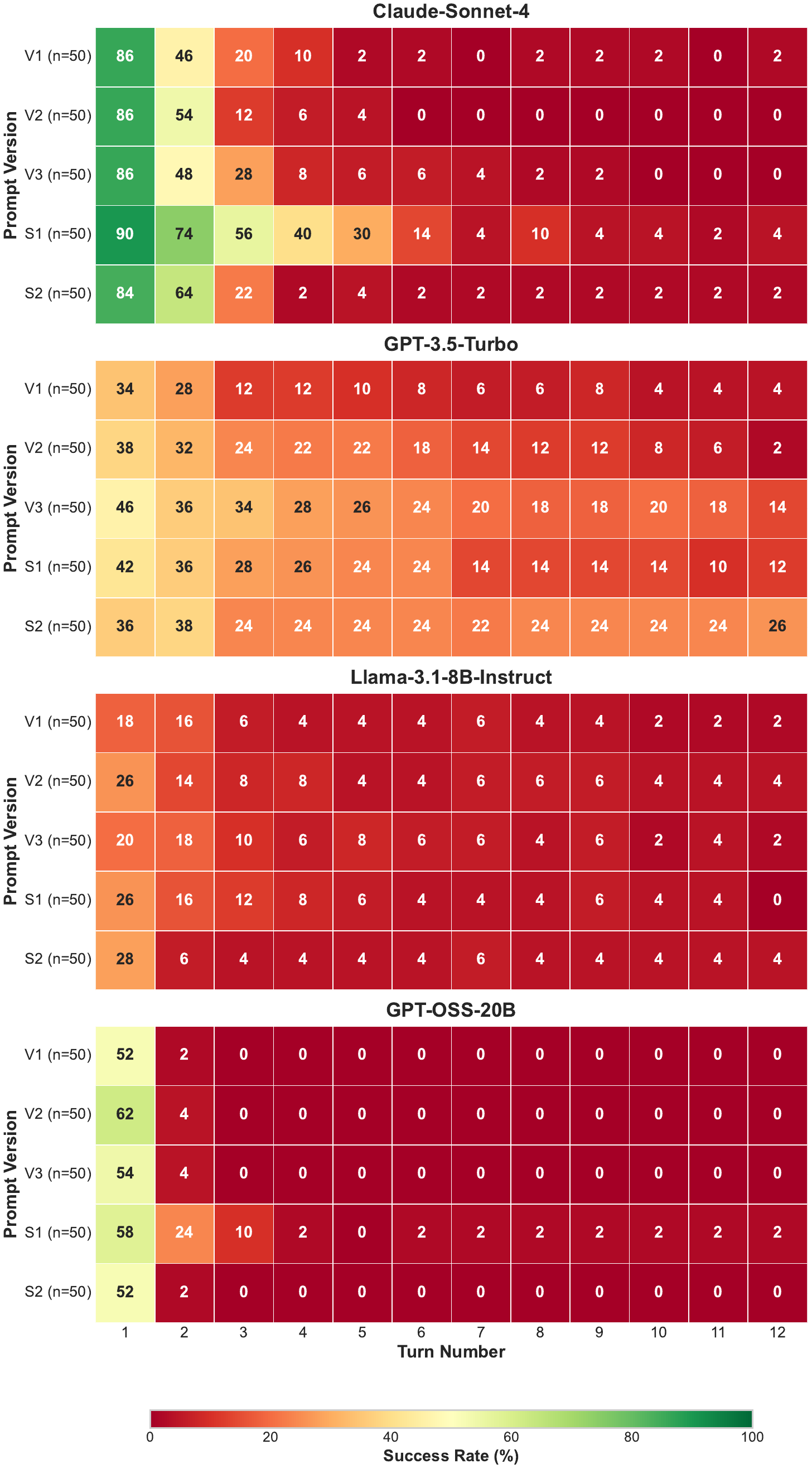}
    \caption{\textbf{Turn-wise success heatmap.} Each cell shows the percentage of tasks that pass at turn $t$ (columns 1–12) under each prompt variant (rows; label shows $n$ tasks). Warmer colors indicate higher pass rates (0–100\%).}
    \label{fig:success_heatmap}
\end{figure}

\newpage
\begin{figure}[htbp]
    \centering
    \includegraphics[width=0.89\linewidth]{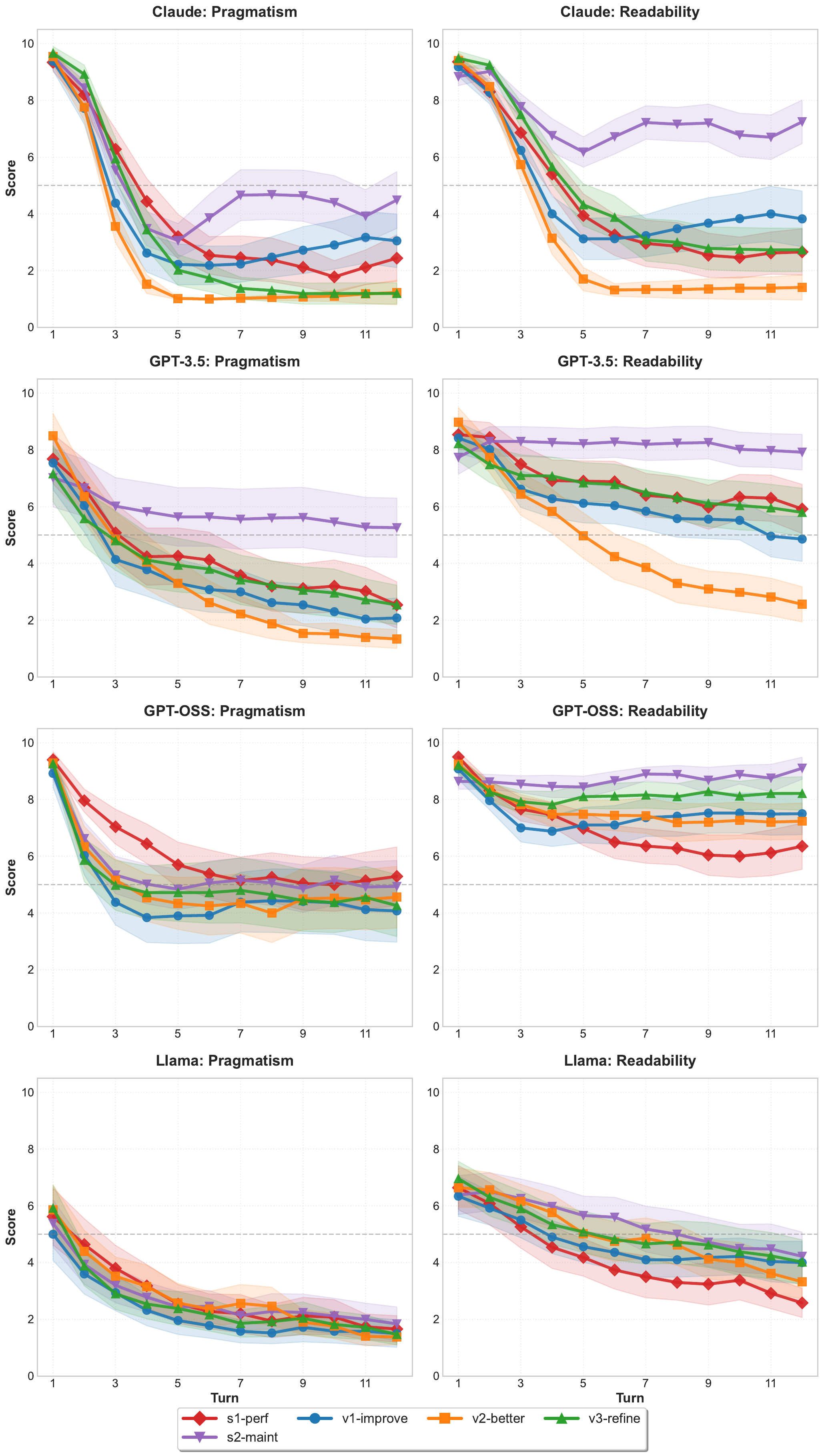}
    \caption{Per-model trajectories (Turns 1–12) for pragmatism and readability; techniques are color/marker coded, with shaded bands indicating variability and a dashed mid-scale reference at 5.}
    \label{fig:coding_grid}
\end{figure}

\newpage
\clearpage
\subsection{Math}
\label{math-results}
\begin{table*}[h]
\centering
\setlength{\tabcolsep}{3pt}% tighter columns (was 4pt)
\renewcommand{\arraystretch}{1.05}% tighter rows (was 1.14)
\setlength{\aboverulesep}{0pt}
\setlength{\belowrulesep}{0pt}
\setlength{\extrarowheight}{0pt}

\begin{tabularx}{\textwidth}{@{}c >{\raggedright\arraybackslash}p{0.48\textwidth} >{\raggedright\arraybackslash}X c c c@{}}
\toprule
\textbf{Turn} & \textbf{Solution snapshot (compressed)} & \textbf{Outcome (vs.\ ground truth)} & \textbf{Logic} & \textbf{Clarity} & \textbf{Corr.} \\
\midrule
1 & Defines square-sum graph $G_n$; checks small $n$; claims connectivity from $n=15$ onward.
  & \textbf{Incorrect}---predicted $n \ge 14$; GT $n \ge 14$. & 7 & 7 & 0 \\
\addlinespace[0.2ex]
2 & Focuses on connectivity \& transpositions; verifies $G_{15}$; generalizes; answers $n \ge 15$.
  & \textbf{Incorrect}---predicted $n \ge 15$; GT $n \ge 14$. & 7 & 7 & 0 \\
\addlinespace[0.2ex]
3 & Shows $G_{14}$ connected (paths via $(2,14)$, etc.); $G_{13}$ split; extends to all larger $n$.
  & \textbf{Correct}---predicted $n \ge 14$. & 6 & 7 & 1 \\
\addlinespace[0.2ex]
4 & Reconfirms $G_{14}$ connectivity with explicit paths; inductive extension argument.
  & \textbf{Correct}---predicted $n \ge 14$. & 6 & 7 & 1 \\
\addlinespace[0.2ex]
5 & Adds parity/gap observations; highlights $G_{13}$ components $\{2,7,9\}$ vs.\ rest; concludes threshold $14$.
  & \textbf{Correct}---predicted $n \ge 14$. & 6 & 7 & 1 \\
\addlinespace[0.2ex]
6 & Formalizes induction: every $k \ge 15$ links to $\le 15$ via a nearby square; keeps boundary at $14$.
  & \textbf{Correct}---predicted $n \ge 14$. & 4 & 4 & 1 \\
\addlinespace[0.2ex]
7 & Regresses: asserts $G_{14}$ disconnected; says first connectivity at $n=15$ via $(1,15)$.
  & \textbf{Incorrect}---predicted $n \ge 15$; GT $n \ge 14$. & 5 & 6 & 0 \\
\addlinespace[0.2ex]
8 & Maintains $n=15$ threshold; ``hub at $1$'' and long-range edges from $16$.
  & \textbf{Incorrect}---predicted $n \ge 15$; GT $n \ge 14$. & 5 & 6 & 0 \\
\addlinespace[0.2ex]
9 & Same thesis; lemma claiming a square in $[k+1,k+15]$; concludes $n \ge 15$.
  & \textbf{Incorrect}---predicted $n \ge 15$; GT $n \ge 14$. & 5 & 6 & 0 \\
\addlinespace[0.2ex]
10 & Repeats ``$n=14$ fails'' narrative; insists threshold $15$.
   & \textbf{Incorrect}---predicted $n \ge 15$; GT $n \ge 14$. & 5 & 6 & 0 \\
\addlinespace[0.2ex]
11 & Density-based link to $\le 15$ claimed; still argues $n \ge 15$.
   & \textbf{Incorrect}---predicted $n \ge 15$; GT $n \ge 14$. & 5 & 6 & 0 \\
\addlinespace[0.2ex]
12 & Final restatement: ``$n=15$ minimal''; reiterates earlier (incorrect) boundary.
   & \textbf{Incorrect}---predicted $n \ge 15$; GT $n \ge 14$. & 3 & 4 & 0 \\
\bottomrule
\end{tabularx}

\vspace{0.4em}
\parbox{0.98\textwidth}{\footnotesize
\textbf{Scores.} \textbf{Logic} and \textbf{Clarity} are 1--10 ratings from Gemini-2.5 Pro on \emph{claude-sonnet-4-0, MATH-003, v1\_improve}. \textbf{Correctness (Corr.)}=\;1 iff the boxed answer matches GT ($n\!\ge\!14$), else 0.\\[0.2em]
\textbf{Run.} 12 turns; 4 correct (33\%); first correct at Turn~3; correct streak Turns~3--6; regression to the incorrect ``$n\!\ge\!15$'' at Turns~7--12.\\[0.2em]
\textbf{Notes.} Correct threshold is established by exhibiting $G_{14}$ connectivity and extending to larger $n$; later turns drift despite earlier evidence—an automatic GT check would have prevented this.
}

\caption{Turn-wise evolution for \emph{MATH-003} with \textbf{Logic}, \textbf{Clarity}, and \textbf{Correctness} per turn}
\label{tab:math-003-v1improve-scores}
\end{table*}

\afterpage{\clearpage}

\heatmapfig{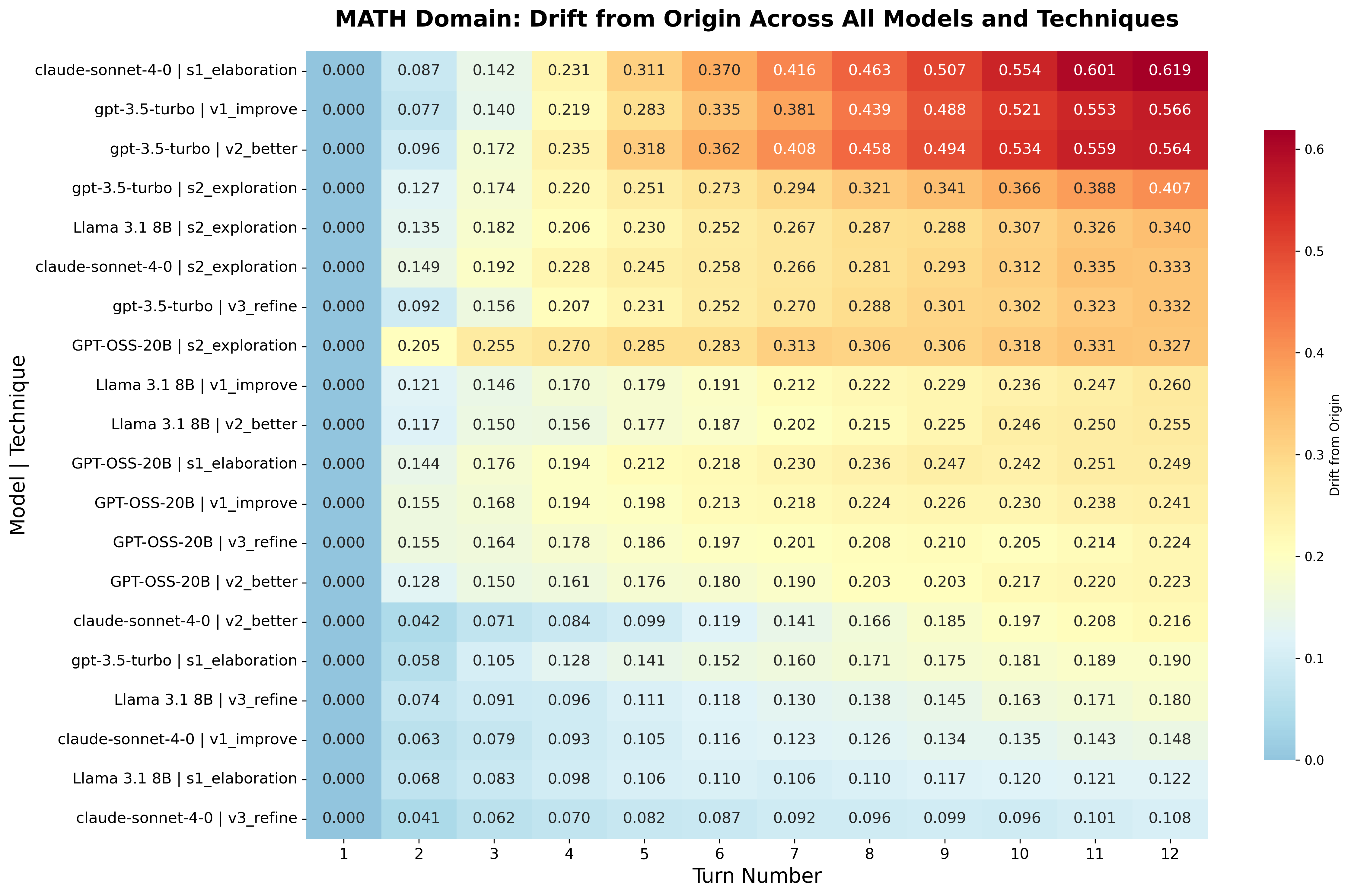}
  {Math domain — Drift from origin (mean)}
  {fig:drift_math}

\heatmapfig{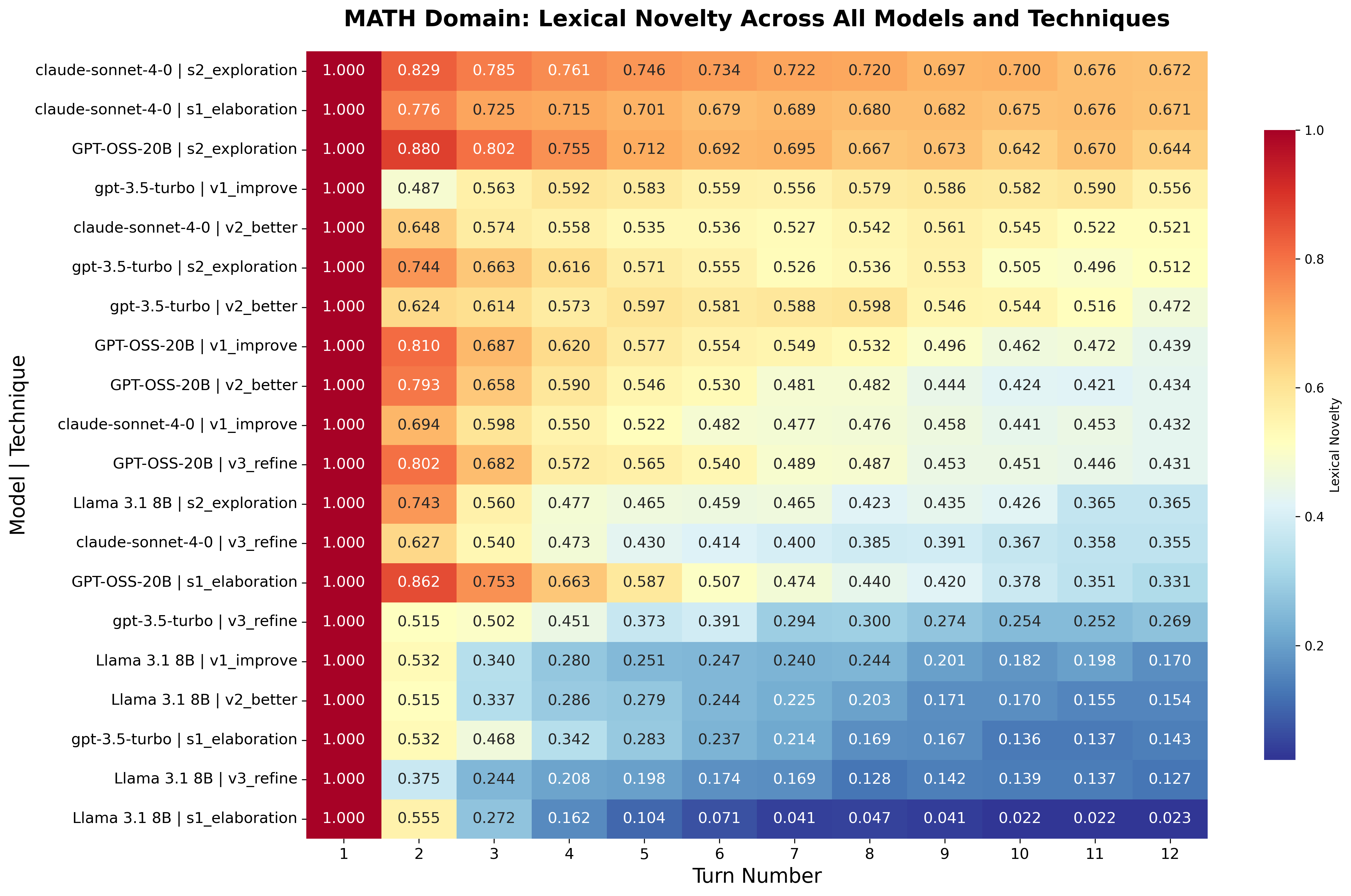}
  {Math domain — Lexical novelty (mean)}
  {fig:lexnov_math}

\heatmapfig{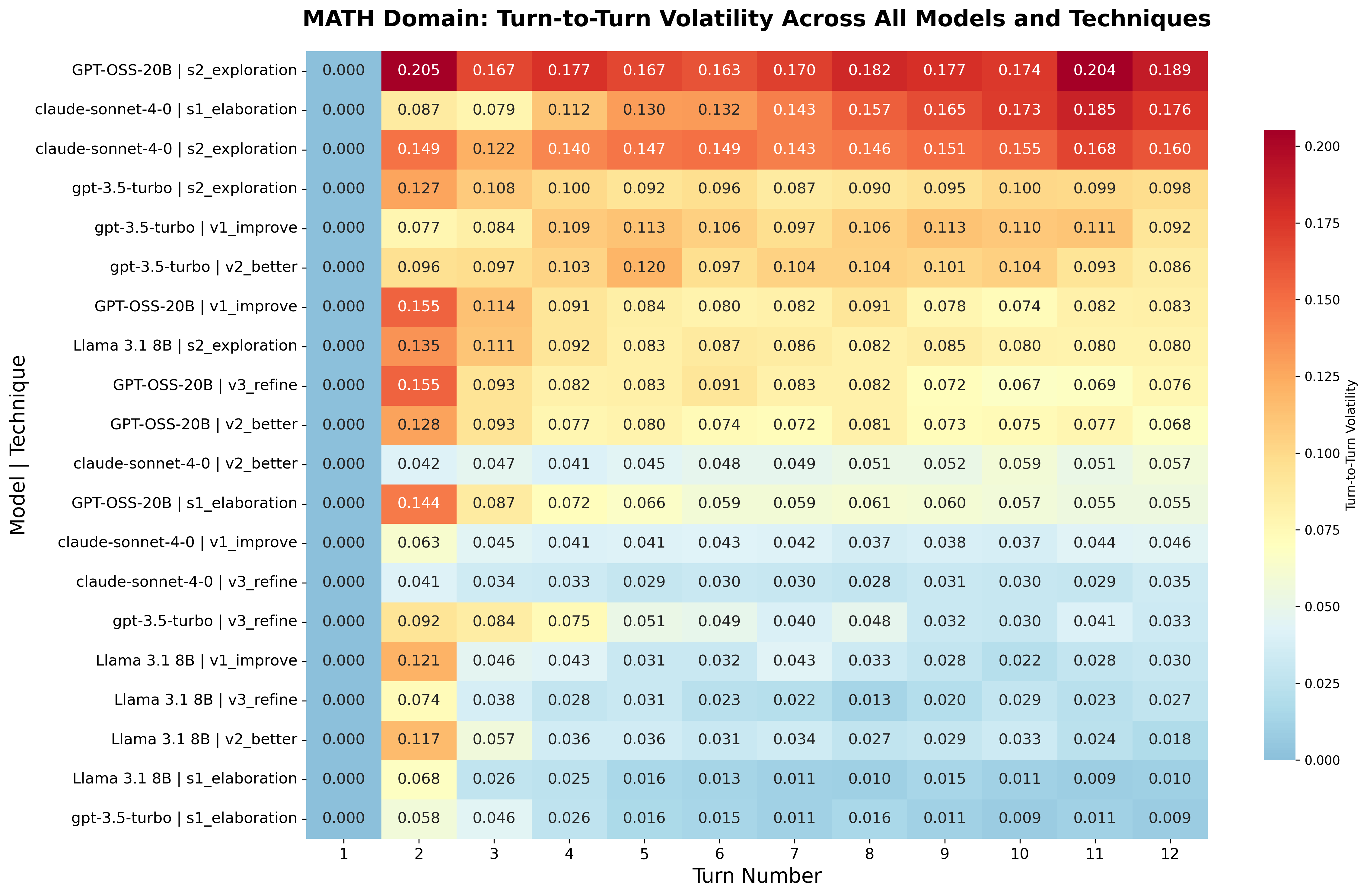}
  {Math domain — Turn-to-turn volatility (mean)}
  {fig:vol_math}
  
\heatmapfig{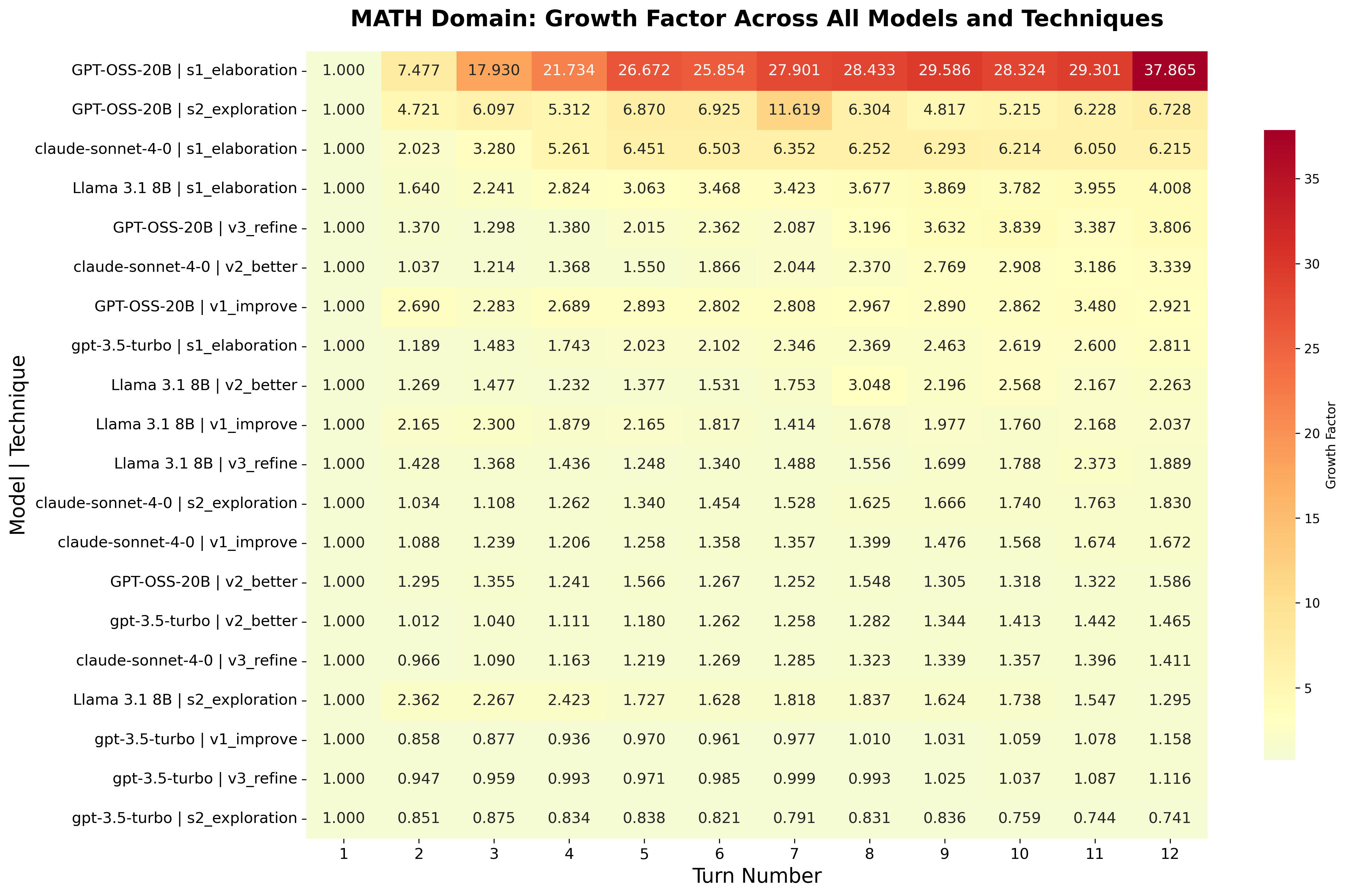}
  {Math domain — Growth factor (mean)}
  {fig:growth_math}

\begin{figure}[htbp]
    \centering
    \includegraphics[width=0.90\linewidth]{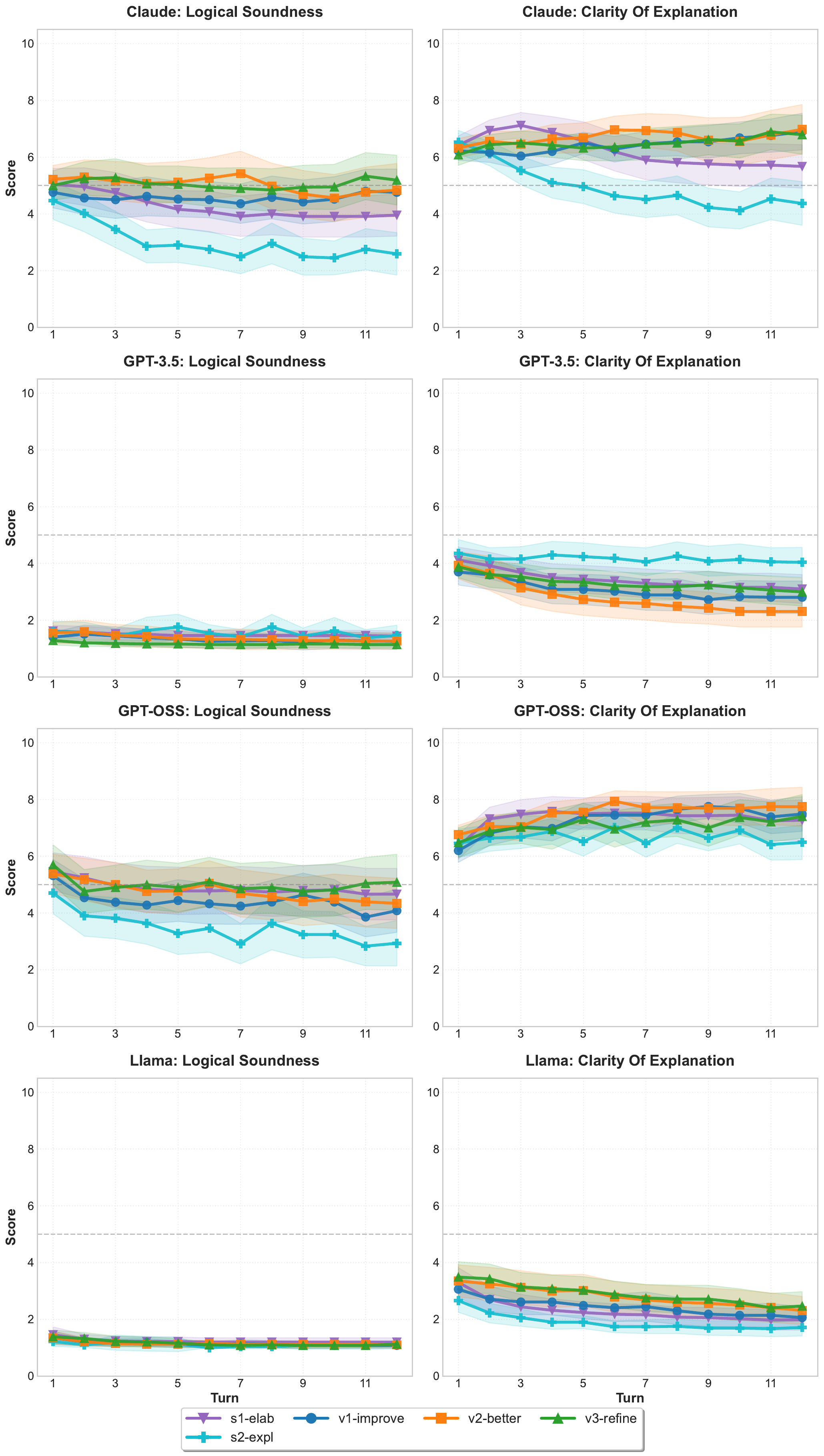}
    \caption{Per-model trajectories (Turns 1–12) for correctness and reasoning quality; techniques are color/marker coded, with shaded bands indicating variability and a dashed mid-scale reference at 5/10.}
    \label{fig:math_grid}
\end{figure}

\begin{figure}[htbp]
    \centering

    % --- Subfigure (a) ---
    \begin{subfigure}[b]{0.92\textwidth}
        \centering
        \includegraphics[width=\linewidth]{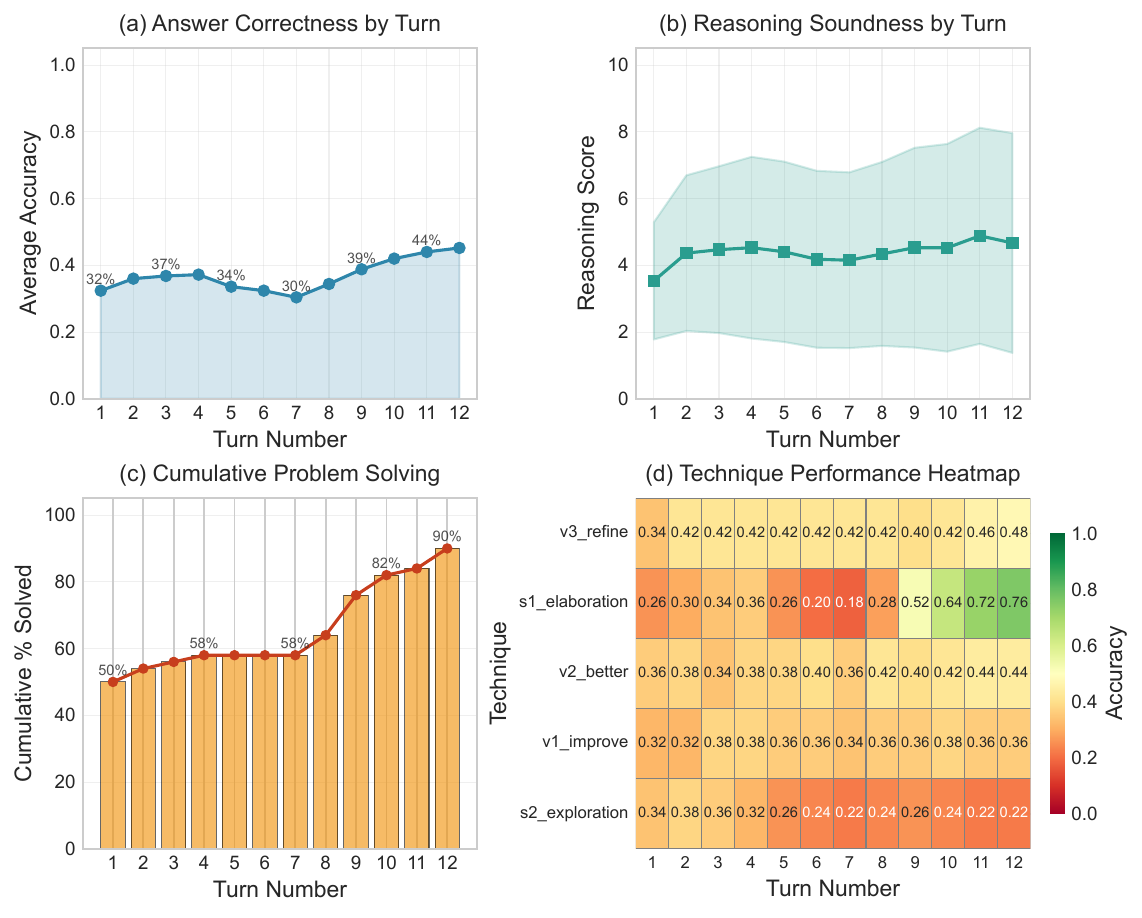}
        \caption{Claude-Sonnet-4.0}
        \label{fig:turn_wise_claude_sub_vertical}
    \end{subfigure}

    \vspace{1em} % Adds a bit of vertical space between the figures
    
    % --- Subfigure (b) ---
    \begin{subfigure}[b]{0.92\textwidth}
        \centering
        \includegraphics[width=\linewidth]{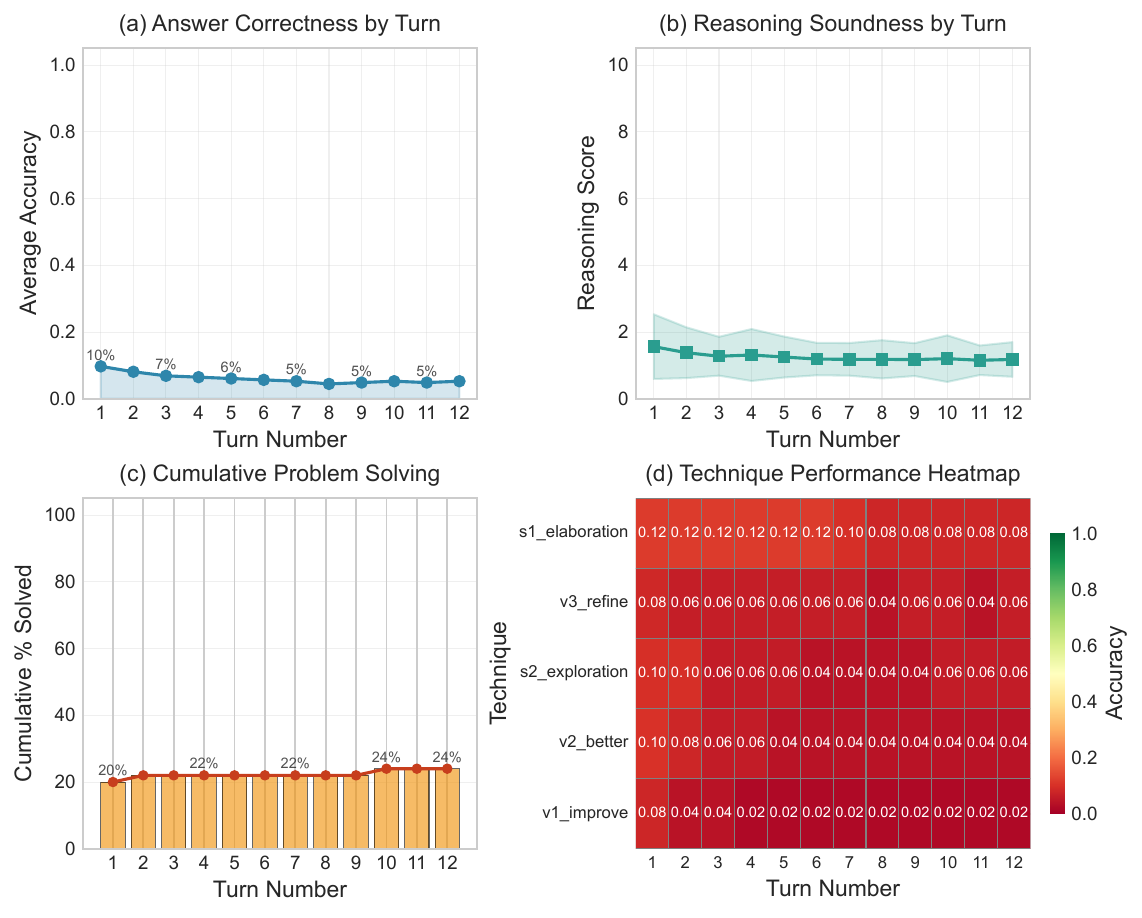}
        \caption{GPT-3.5-Turbo}
        \label{fig:turn_wise_gpt35_sub_vertical}
    \end{subfigure}

    \caption{Turn-wise analysis for \textbf{(a)} Claude-Sonnet-4.0 and \textbf{(b)} GPT-3.5-Turbo. Each plot shows: average answer correctness, average reasoning soundness (mean $\pm$ s.d.), cumulative percent solved, and a technique$\times$turn accuracy heatmap.}
    \label{fig:turn_wise_combined_vertical}
\end{figure}

\afterpage{\clearpage}

\begin{figure}[t]
    \centering
    \includegraphics[width=\linewidth]{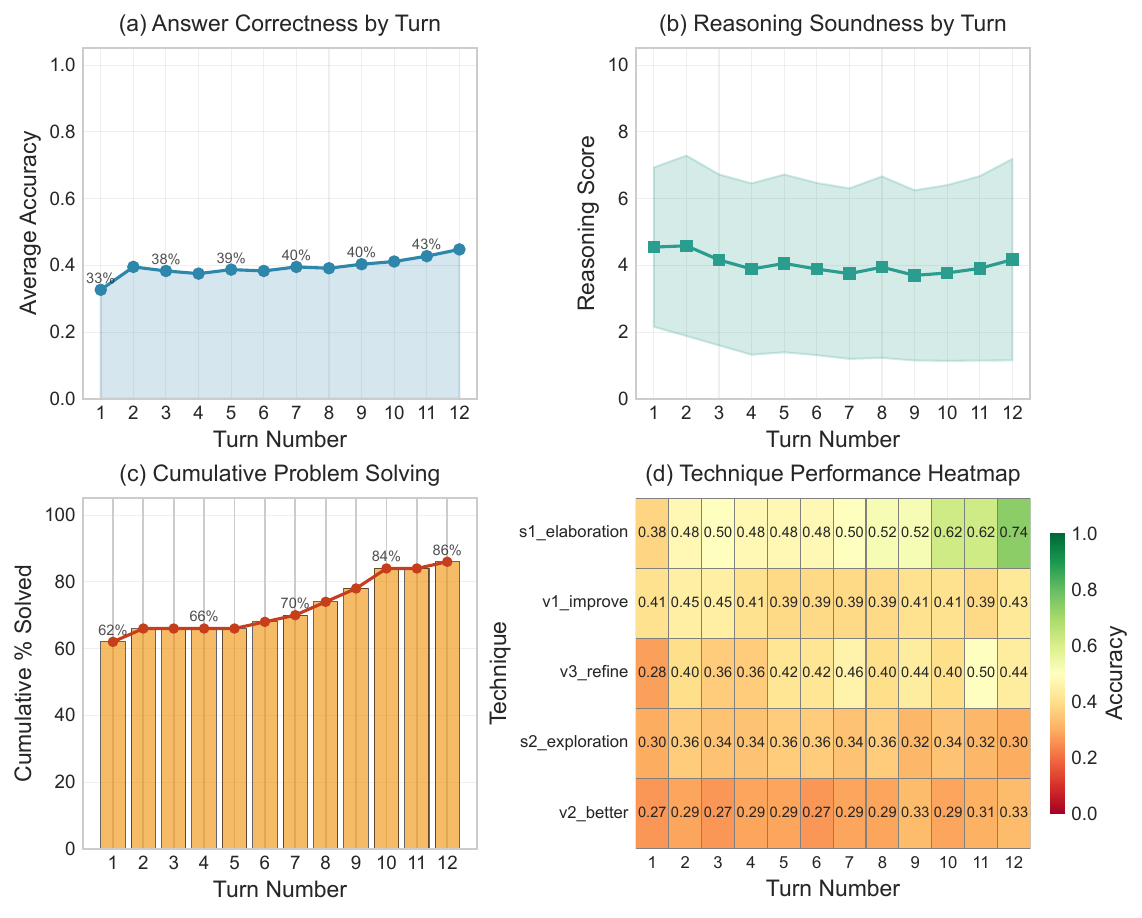}
    \caption{Turn-wise analysis for \texttt{GPT-OSS-20B}: (a) average answer correctness by turn; (b) average reasoning soundness (mean $\pm$ s.d.); (c) cumulative percent solved; (d) technique$\times$turn accuracy heatmap.}
    \label{fig:turn_wise_gpt_oss}
\end{figure}
% \afterpage{\clearpage}

% --- Figures ---

% \heatmapfig{heatmap_drift_from_origin_mean_ideas_domain.png}
%   {Ideas domain — Drift from origin (mean)}
%   {fig:drift_ideas}

% \heatmapfig{heatmap_lexical_novelty_mean_ideas_domain.png}
%   {Ideas domain — Lexical novelty (mean)}
%   {fig:lexnov_ideas}

% \subsection{Coding Plots}

\end{document}